\documentclass[times, review, 10pt]{elsarticle}

\usepackage{graphicx} 
\usepackage{multirow}
\usepackage{tabularx}
\usepackage{booktabs}

\journal{Journal of \LaTeX\ Templates}

\newcounter{example}

\usepackage{afterpage}
\usepackage{capt-of}
\usepackage{xcolor}

\begin{document}

\begin{frontmatter}

\title{Comparative Evaluation of ChatGPT and DeepSeek Across Key NLP Tasks: Strengths, Weaknesses, and Domain-Specific Performance}

\author{Wael Etaiwi}
\ead{w.etaiwi@psut.edu.jo}

\author{Bushra Alhijawi}
\ead{b.alhijawi@psut.edu.jo}

\address{Princess Sumaya University for Technology, Amman, Jordan}

\begin{abstract}
 The increasing use of large language models (LLMs) in natural language processing (NLP) tasks has sparked significant interest in evaluating their effectiveness across diverse applications. While models like ChatGPT and DeepSeek have shown strong results in many NLP domains, a comprehensive evaluation is needed to understand their strengths, weaknesses, and domain-specific abilities. This is critical as these models are applied to various tasks, from sentiment analysis to more nuanced tasks like textual entailment and translation. This study aims to evaluate ChatGPT and DeepSeek across five key NLP tasks: sentiment analysis, topic classification, text summarization, machine translation, and textual entailment. A structured experimental protocol is used to ensure fairness and minimize variability. Both models are tested with identical, neutral prompts and evaluated on two benchmark datasets per task, covering domains like news, reviews, and formal/informal texts. The results show that DeepSeek excels in classification stability and logical reasoning, while ChatGPT performs better in tasks requiring nuanced understanding and flexibility. These findings provide valuable insights for selecting the appropriate LLM based on task requirements.

\end{abstract}


\end{frontmatter}

\section{Introduction}
\label{sec:introduction}

{Artificial} Intelligence (AI) technology has significantly transformed various domains, including healthcare, finance, education, and scientific research, by enabling machines to perform complex cognitive tasks \cite{wang2025empowering}. AI-driven solutions have progressed from rule-based and statistical models to deep learning approaches, demonstrating remarkable computer vision, speech processing, and natural language processing (NLP) capabilities \cite{chen2024artificial}. 

A breakthrough in AI research has been the development of generative AI (Gen AI) \cite{feuerriegel2024generative}. Gen AI is a class of machine learning models designed to create new content rather than solely analyze or classify existing data \cite{feuerriegel2024generative}. Unlike traditional AI models that rely on explicit rules and predefined patterns, Gen AI models utilize deep neural networks to generate text, images, audio, and code with contextually appropriate outputs \cite{feuerriegel2024generative}. These models have applications in machine translation, content generation, code synthesis, and automated reasoning. Gen AI has demonstrated remarkable capabilities in storytelling \cite{yoo2024leveraging}, dialogue generation \cite{liu2024speak}, content summarization \cite{van2024field}, and question answering \cite{li2024flexkbqa}. Large Language Models (LLMs) enable these capabilities by earning from vast text corpora and then being fine-tuned for specific tasks. LLMs have become the foundation of modern NLP applications due to their ability to process and generate human-like text with contextual understanding \cite{kumar2024large}. These models leverage self-attention mechanisms to capture long-range dependencies in text, enabling them to perform a wide range of linguistic tasks \cite{islam2023comprehensive}. Recent LLM advancements have resulted in highly sophisticated models, such as ChatGPT \cite{achiam2023gpt} and DeepSeek \cite{liu2024deepseek}, demonstrating state-of-the-art performance across multiple NLP benchmarks. ChatGPT, developed by OpenAI, is a transformer-based model designed for various language generation tasks, including dialogue systems, text summarization, code generation, and question answering \cite{achiam2023gpt}. It has been fine-tuned with reinforcement learning from human feedback to enhance its conversational abilities, coherence, and contextual understanding. ChatGPT has gained widespread adoption due to its ability to generate human-like responses, making it a leading choice for chatbot applications, virtual assistants, and creative writing tasks \cite{bahrini2023chatgpt}. In contrast, DeepSeek, developed by DeepSeek AI, is an advanced LLM designed to focus on multilingual processing, domain-specific applications, and knowledge-intensive tasks \cite{liao2025deepseek}. It leverages a highly optimized training pipeline and extensive datasets to improve factual accuracy, logical reasoning, and robustness across diverse NLP benchmarks. While both models are built on the transformer architecture, their training methodologies, optimization strategies, and application focus differ, necessitating a detailed comparative analysis to evaluate their strengths, weaknesses, and suitability for various NLP tasks.

Many studies evaluated the performance of ChatGPT and DeepSeek in different applications. However, to the best of our knowledge, few studies have focused on evaluating and comparing ChatGPT and DeepSeek on NLP tasks using benchmark datasets. Fu et al. \cite{fu2024can}  investigated whether ChatGPT can effectively evaluate the quality of planning documents, specifically in the context of climate change plans. By comparing ChatGPT's evaluations with those of human coders, the research reveals that ChatGPT's results align reasonably well with traditional methods, although it struggles with planning-specific jargon. Thelwall \cite{thelwall2024can} evaluated ChatGPT 4.0's ability to assess journal article quality using UK Research Excellence Framework 2021 guidelines, applied to 51 of the author's articles. Also, the author compared ChatGPT results with self-evaluations. Caramancion \cite{caramancion2023news} evaluated the capability of ChatGPT 3.5 and 4.0, Google's Bard/LaMDA, and Microsoft's Bing AI in verifying the truthfulness of news items using black box testing. Johnson et al. \cite{johnson2023assessing} assessed the accuracy and completeness of ChatGPT in responding to 284 medical questions created by 33 physicians across 17 specialties. They analyzed its performance using Likert scale ratings and statistical comparisons, finding generally high accuracy with limitations in complex cases, highlighting the need for further model refinement. Antaki et al. \cite{antaki2023evaluating}  assessed ChatGPT's accuracy in answering ophthalmology exam questions, showing improved performance with the Plus version and greater consistency across topics. Dunder et al. \cite{dunder2024kattis} evaluated the capability of ChatGPT in solving programming tasks using Kattis, revealing proficiency in simple problems but challenges with complex ones. ChatGPT successfully solved 19 out of 127 problems. Elyoseph et al. \cite{elyoseph2023chatgpt} evaluated ChatGPT's capacity to identify and describe emotions using an objective emotional awareness scale, demonstrating superior performance with progressive improvement over time. The results indicate potential applications in cognitive training and psychiatric assessment. Khlaif et al. \cite{khlaif2023potential} assessed ChatGPT's role in scientific research, showcasing its ability to produce high-quality content but identifying challenges in research methodology and literature review. Deandres et al. \cite{deandres2024good} investigated ChatGPT's potential in face biometrics, including tasks like face verification and soft-biometrics estimation, revealing its ability to enhance the explainability of automatic decisions. The results demonstrate the effectiveness of LLMs such as ChatGPT in improving transparency and robustness in human-related biometric systems. Table \ref{RWSum} provides an overview of the experimental evaluation of ChatGPT across several tasks.

\begin{table}[h!]
\caption{Summary of the Experimental Evaluation of ChatGPT on Various Tasks}
\label{RWSum}
\hspace*{-2.5cm}
\resizebox{6.5in}{!}{
\begin{tabular}{|c|c|l|l|}
\hline
\textbf{Article}     & \textbf{Year} & \multicolumn{1}{c|}{\textbf{Objective}} & \multicolumn{1}{c|}{\textbf{Description}}                                                                                                                                                                                                                                                                                                                                      \\ \hline
\cite{fu2024can}            & 2024          & ChatGPT vs Human                        & \begin{tabular}[c]{@{}l@{}}- Explored if ChatGPT can evaluate planning documents, focusing on climate change plans.\\ - Compared ChatGPT’s evaluation results with those from human coders.\end{tabular}                                                                                                                                                                       \\ \hline
\cite{thelwall2024can}      & 2024          & Evaluation                              & \begin{tabular}[c]{@{}l@{}}- Assessed whether ChatGPT 4.0 can automate research evaluations of journal articles, focusing\\ on its accuracy.\\ - Tested ChatGPT 4.0’s ability to evaluate journal articles using the UK Research Excellence \\ Framework 2021 guidelines.\\ - Compared ChatGPT-4's evaluations with the author’s self-assessments of 51 articles.\end{tabular} \\ \hline

\cite{dunder2024kattis}     & 2024          & Evaluation                              & \begin{tabular}[c]{@{}l@{}}- Evaluated ChatGPT’s ability to generate code solutions for programming tasks in introductory \\ computer science courses.\\ - Tested ChatGPT on 127 randomly selected programming problems from Kattis, an automated \\ grading tool.\end{tabular}                                                                                                \\ \hline
\cite{deandres2024good}     & 2024          & ChatGPT vs Other                        & \begin{tabular}[c]{@{}l@{}}- Explored ChatGPT 4.0's capability in face biometrics tasks, including face verification, \\ soft-biometrics estimation, and result explainability.\\ - Experiments using public benchmarks were conducted to assess ChatGPT's performance in \\ face biometrics and compare it with state-of-the-art methods.\end{tabular}                        \\ \hline
\cite{caramancion2023news}  & 2023          & ChatGPT vs Other                        & \begin{tabular}[c]{@{}l@{}}- Assessed the proficiency of popular LLMs in verifying the truthfulness of news items using \\ black box testing.\\ - Evaluated ChatGPT 3.5 \& 4.0, Google's Bard/LaMDA, and Microsoft's Bing AI on 100 \\ fact-checked news items.\end{tabular}                                                                                                   \\ \hline
\cite{johnson2023assessing} & 2023          & Evaluation                              & \begin{tabular}[c]{@{}l@{}}- Evaluated the accuracy and completeness of ChatGPT’s responses to medical queries across \\ various specialties.\\ - Assessed ChatGPT’s performance on 284 medical questions created by 33 physicians from \\ 17 specialties.\end{tabular}                                                                                                        \\ \hline
\cite{antaki2023evaluating} & 2023          & Evaluation                              & \begin{tabular}[c]{@{}l@{}}- Evaluated ChatGPT's accuracy in answering ophthalmology-related multiple-choice questions.\\ - Tested ChatGPT on two question banks (BCSC and OphthoQuestions) used for the OKAP exam.\end{tabular}                                                                                                                                               \\ \hline
\cite{elyoseph2023chatgpt}  & 2023          & ChatGPT vs Human                        & \begin{tabular}[c]{@{}l@{}}- Investigated ChatGPT’s ability to identify and describe emotions using the levels of emotional \\ awareness scale.\\ - Analyzed ChatGPT’s responses to 20 emotional scenarios and compared its performance to \\ humans.\end{tabular}                                                                                                             \\ \hline
\cite{khlaif2023potential}  & 2023          & Evaluation                              & \begin{tabular}[c]{@{}l@{}}- Examined ChatGPT's potential in research by assessing the quality of AI-generated articles, \\ data analysis, and literature reviews.\\ - Generated four articles and 50 abstracts using ChatGPT, evaluated by 23 reviewers, and \\ analyzed using ANOVA and thematic analysis.\end{tabular}                                                      \\ \hline
\end{tabular}}
\end{table}

In that context, this study aims to evaluate the performance of ChatGPT and DeepSeek across five core NLP applications: sentiment analysis, topic classification, text summarization, machine translation, and textual entailment in the English language. For each application, two benchmark datasets are selected for comprehensive evaluation. The performance of both models is then compared using an offline evaluation methodology. To summarize the main contribution of this research paper: 

\begin{itemize}
\item  Evaluating the performance of ChatGPT on sentiment analysis, topic classification, text summarization, machine translation, and textual entailment tasks in the English language.

\item  Assessing the performance of DeepSeek on sentiment analysis, topic classification, text summarization, machine translation, and textual entailment tasks in the English language. 

\item Comparing the performance of ChatGPT and DeepSeek based on offline evaluation methodology.

\end{itemize}

The remainder of this paper is organized as follows. The next section provides an overview of ChatGPT and DeepSeek. Section \ref{method} outlines the methodology employed in this study to perform an offline evaluation of both models. Section \ref{res} presents, compares, and discusses the results obtained from the two LLMs. Finally, Section \ref{con} concludes the paper and suggests potential directions for future research.




\section{ChatGPT and DeepSeek Overview}

Recent LLM advancements have led to the development of sophisticated systems such as OpenAI's ChatGPT and DeepSeek's models, significantly enhancing NLP capabilities. These models exhibit impressive performance across various tasks, including question-answering, text generation, and summarization.

ChatGPT, developed by OpenAI\footnote{https://openai.com/}, is based on the GPT-4 architecture, a dense transformer model trained on extensive internet text. It has been fine-tuned using Reinforcement Learning with Human Feedback, improving alignment with user intent and safety constraints \cite{Ouyang2022}. This fine-tuning process enables ChatGPT to excel in general-purpose reasoning, complex problem-solving, and creative tasks such as code generation \cite{Coello2024} and academic writing \cite{Rezaei2024}. OpenAI has introduced multimodal capabilities in ChatGPT, enabling it to process text and images and enhancing its versatility in generating lifelike images and coherent text. Additionally, OpenAI has launched subscription services like DALL-E and ChatGPT Plus, offering users access to advanced features and models, including GPT-4.5, which provides up-to-date information retrieval and supports file and image uploads \cite{Gonzlez2024}.

DeepSeek\footnote{https://www.deepseek.com/} has adopted a Sparse Mixture of Experts architecture, activating only a subset of model parameters during inference to optimize computational efficiency \cite{Liu2024}. The latest model, DeepSeek-V3, released in March 2025, boasts 671 billion parameters with 37 billion activated per token, demonstrating significant improvements in reasoning and coding capabilities compared to its predecessors. DeepSeek's models have shown efficiency in running on consumer-grade hardware, with DeepSeek-V3 achieving 20 tokens per second on a Mac Studio, challenging traditional notions about AI model deployment requirements  \cite{Jiang2025}. Furthermore, DeepSeek has narrowed the AI development gap with leading U.S. companies, with experts suggesting a convergence within three months in certain areas \cite{Mo2025DeepSeek}.

\section{Methodology}
\label{method}

This section outlines the methodology followed in evaluating the performance of ChatGPT and DeepSeek across five NLP tasks: sentiment analysis, topic classification, text summarization, machine translation, and textual entailment. The evaluation is conducted using an offline evaluation methodology, ensuring a standardized comparison. The methodology includes details on experimental setup (Section \ref{expS}), datasets (Section \ref{data}), response collection (Section \ref{Resp}), and evaluation measures (Section \ref{evalM}).

\subsection{Experimental Setup and Protocol}
\label{expS}

A well-defined experimental protocol is established to ensure a systematic and unbiased evaluation of ChatGPT and DeepSeek. This protocol governs the querying process, response collection, and evaluation framework, minimizing sources of variability and ensuring a fair comparison. The key components of the experimental design are detailed as follows:

\begin{itemize}
\item Prompt standardization: A set of predefined, structured prompts is designed for each NLP task to ensure consistency in model inputs. The prompts are crafted to be neutral, unambiguous, and representative of real-world use cases, minimizing biases that may affect model responses. The following prompt standardization principles are applied to eliminate inconsistencies in input structure:

\begin{itemize}
\item Uniformity across models: Identical prompts are used for ChatGPT and DeepSeek across all tasks to ensure a fair comparison. These prompts are designed to be clear, concise, and unambiguous, adhering to best practices in LLM prompt engineering. Additionally, no extra context or system instructions are provided beyond what is essential for task completion.

\item Example-based prompting (Few-Shot Learning): For textual entailment and machine translation tasks, few-shot examples are tested to assess the impact of context on response accuracy. Example-based prompting is applied consistently across models to ensure fairness and prevent bias.
\end{itemize}

\item Dataset splitting and consistency: To ensure a diverse and comprehensive evaluation, two benchmark datasets are selected for each NLP task, with both models evaluated on identical dataset splits to ensure fairness. The LLMs are assessed on balanced class distributions to prevent bias. Additionally, text examples from various domains, including news, reviews, and formal and informal texts, are incorporated to evaluate model generalization.

\end{itemize}

The considered structured protocol ensures robust and reliable performance evaluation, minimizing potential biases or inconsistencies.

\subsection{NLP Tasks and Benchmark Datasets}
\label{data}
 
Two benchmark datasets are selected to assess the performance of ChatGPT and DeepSeek across five core NLP tasks: sentiment analysis, topic classification, text summarization, machine translation, and textual entailment in the English language. These datasets, which are chosen to provide a comprehensive evaluation of the models' capabilities, are summarized in Table \ref{datasetSummary}.

\begin{table}[h!]
\caption{Summary of Benchmark Datasets Used in the Experiments}
\label{datasetSummary}
\hspace*{-2cm}
\resizebox{7in}{!}{
\begin{tabular}{|l|l|l|c|c|}
\hline
Task & Dataset Name & Domain & Count of Records & Count of Testing Samples \\ \hline
\multirow{2}{*}{Sentiment analysis}   & Multilingual Sentiment Datasets \cite{augenstein-etal-2017-semeval} & Scientific Documents & 3036 & 150 \\ \cline{2-5} 
     & IMDB \cite{maasIMDB2011} & Movie Reviews & 25000 & 100 \\ \hline

\multirow{2}{*}{Topic classification} & AG News Classification Dataset \cite{Corso2005} & News Articles & 1 million & 80 \\ \cline{2-5} 
  & Web of Science Dataset \cite{hKowsari2018,kowsari2017HDLTex} & Scientific Articles & 64688 & 100 \\ \hline

\multirow{2}{*}{Text Summarization}   & CNN / Daily Mail Summarization Dataset \cite{see-etal-2017-get} & News Articles & 300000 & 50 \\ \cline{2-5} 
     & Gigaword \cite{RushGigaword2024} & News Articles & 4 millions & 50 \\ \hline

\multirow{2}{*}{Machine Translation}  & AraBench \cite{Sajjad2020} & Diverse textual genres & 173000 & 50 \\ \cline{2-5} 
   & ArzEn-MultiGenre \cite{AlSabbagh2024} & Novels, subtitles, and songs & 25557 & 50 \\ \hline

\multirow{2}{*}{Textual entailment} & Scitail \cite{Khot2018} & - & 27026 & 50\\ \cline{2-5} 
  & FraCaS \cite{cooper1996using} & -  & 350 & 75\\ \hline
\end{tabular}}
\end{table}

\subsubsection{Sentiment Analysis Datasets}

Sentiment analysis involves analyzing textual data, such as comments, reviews, and opinions, to determine the sentiment expressed toward a specific subject, such as a product, event, or individual \cite{Biltawi7476075}. The performance of ChatGPT and DeepSeek in sentiment classification is evaluated using the IMDB and Multilingual Sentiment datasets. The IMDB dataset \footnote{https://github.com/Ankit152/IMDB-sentiment-analysis} is designed for binary sentiment classification. This dataset consists of 25,000 movie reviews for training and an additional 25,000 reviews for testing \cite{maasIMDB2011}. The Multilingual Sentiment dataset \footnote{https://github.com/tyqiangz/multilingual-sentiment-datasets} combines several datasets to capture sentiment across multiple languages, including various Asian languages. The English portion of this dataset, sourced from the SemEval dataset \cite{augenstein-etal-2017-semeval}, includes 1,840 training, 871 testing, and 325 validation records. Each record is classified as positive, neutral, or negative. Random samples of 50 instances per class are selected as testing samples for the experiments to ensure a consistent and fair evaluation.

\subsubsection{Topic Classification Datasets}

Topic classification is a text-mining technique that categorizes documents into predefined categories or labels based on their content \cite{Sunagar2021}. In this study, ChatGPT and DeepSeek’s performance in topic classification is evaluated using two benchmark datasets: the AG News Classification dataset \cite{Corso2005} and the Web of Science dataset \cite{hKowsari2018,kowsari2017HDLTex}. The AG News dataset \footnote{$http://groups.di.unipi.it/gulli/AG_corpus_of_news_articles.html$} consists of over one million news articles collected from more than 2,000 sources over a year, categorized into four main classes: Business, Sci/Tech, Sports, and World. For the AG News dataset, 20 testing samples are randomly selected per class. The Web of Science topic classification dataset \footnote{https://data.mendeley.com/datasets/9rw3vkcfy4/6} serves as a standardized benchmark for text classification, focused explicitly on topic categorization in scientific literature. This dataset includes research article abstracts assigned to seven predefined subject areas: Computer Science (CS), Electrical and Computer Engineering (ECE), Psychology, Mechanical and Aerospace Engineering (MAE), Civil Engineering, Medical Science, and Biochemistry. It is available in three versions: WOS-11967, WOS-46985, and WOS-5736, containing 11,967, 46,985, and 5,736 abstracts, respectively. For the Web of Science dataset, 100 testing samples are randomly selected, with varying instances per class.

\subsubsection{Text Summarization Datasets}

Text summarization is the automated process of generating a concise and coherent summary from a given text \cite{silva2015automatic}. It can be classified into two main types: extractive and abstractive summarization. Extractive summarization involves selecting key sentences or paragraphs directly from the source text using scoring algorithms to identify the most relevant content. In contrast, abstractive summarization generates new sentences based on a semantic understanding of the document, producing summaries that may not include the original text verbatim but retain its essential meaning.

In this study, two benchmark datasets are used to evaluate the performance of ChatGPT and DeepSeek in text summarization: the CNN/Daily Mail Summarization Dataset \cite{see-etal-2017-get} and the Gigaword Dataset \cite{RushGigaword2024}. The CNN/DailyMail dataset \footnote{https://cs.nyu.edu/~kcho/DMQA/} is an English-language corpus containing over 300,000 news articles from CNN and the Daily Mail. It is widely used for evaluating text summarization models. The Gigaword text summarization dataset \footnote{https://catalog.ldc.upenn.edu/LDC2012T21} is a large-scale dataset derived from the English Gigaword corpus, consisting of millions of news articles from respected sources like The New York Times, Associated Press, and Xinhua News Agency. This dataset is extensively used for both text summarization and headline generation tasks.
For the evaluation, 100 random instances from each dataset are selected as testing samples.

\subsubsection{Machine Translation Dataset}

Machine Translation is the automatic conversion of text from one language to another using computational models and techniques \cite{Wang2022}. It is widely used in applications like online translation services (e.g., Google Translate and DeepL) and real-time speech translation.

To evaluate the performance of ChatGPT and DeepSeek in machine translation, this study employs two benchmark datasets: AraBench \cite{Sajjad2020} and the ArzEn-MultiGenre \cite{AlSabbagh2024} datasets. The AraBench dataset \footnote{https://alt.qcri.org/resources1/mt/arabench/} is a dialectal Arabic-to-English machine translation corpus. It contains approximately 173,000 sentences categorized into 4 coarse-grained, 15 fine-grained, and 25 city-level classifications. The dataset includes a variety of textual genres such as media, chat, religion, and travel, each with varying levels of dialectal complexity. The ArzEn-MultiGenre dataset \footnote{https://data.mendeley.com/datasets/6k97jty9xg/5} is a manually translated parallel corpus consisting of 25,557 sentence pairs in Egyptian Arabic and English. It covers three genres: novels, subtitles, and songs.
For the evaluation process, 100 random samples are selected as testing data from each dataset to ensure a consistent and reliable performance comparison between the two models.

\subsubsection{Textual Entailment Dataset}

Textual entailment refers to the semantic relationship between two text fragments, where one fragment logically follows or is implied by the other \cite{Fidelangeli2025}. This relationship indicates that the meaning of one text can be inferred from the other, establishing a form of semantic implication. Automating textual entailment recognition is crucial for enhancing various NLP tasks, including information retrieval, extraction, question answering, text summarization, and machine translation.

To evaluate the performance of ChatGPT and DeepSeek in textual entailment recognition, this study utilizes SciTail \cite{Khot2018} and FraCaS dataset \cite{cooper1996using}. The SciTail dataset \footnote{https://leaderboard.allenai.org/scitail/submissions/get-started} is derived from multiple-choice science exams and web sentences. It contains 27,026 examples, with 10,101 labeled as "entails" and 16,925 labeled as "neutral." Each example consists of a question, and the correct answer choice is transformed into a hypothesis, with relevant text retrieved from a vast corpus of web sentences as the premise. The $FraCaS$ dataset \footnote{https://nlp.stanford.edu/~wcmac/downloads/fracas.xml} is an inference test suite created to evaluate the inferential capabilities of various NLP systems. It consists of approximately 350 labeled examples, with annotations to assess the recognition of entailment, contradiction, and neutrality. For evaluation, 50 samples per class are randomly selected as testing data, ensuring a fair and consistent performance comparison between the models.

\subsection{LLM Querying and Response Collection}
\label{Resp}

To evaluate the performance of ChatGPT and DeepSeek, we interact with both models directly through their user interfaces. This approach ensures that the study reflects real-world usage scenarios where users manually input queries and receive responses without API-based automation. Querying follows a structured, consistent methodology to ensure fairness and comparability across tasks. The response collection procedure includes the following steps:

\begin{itemize}
\item Consistent input formatting: The same prompts are used for both models to ensure a fair comparison. They are carefully designed to be clear, unambiguous, and aligned with best practices in prompt engineering. No additional instructions or system messages are provided beyond those necessary for the task. Table \ref{examplePrompts} provides examples of the prompts employed for each NLP task.

\begin{table}[h!]
\caption{Sample Queries for NLP Tasks}
\label{examplePrompts}

\begin{tabular}{|c|l|}
\hline
\textbf{Task}                                                  & \multicolumn{1}{c|}{\textbf{Sample Query}}                                                                                                                                                                                                                                                         \\ \hline
\begin{tabular}[c]{@{}c@{}}Sentiment\\ Analysis\end{tabular}   & \begin{tabular}[c]{@{}l@{}}Determine the sentiment of the following review, it \\ should be either positive, neutral, or negative: 'The \\ movie was fantastic! The storyline was engaging, \\ and the acting was top-notch.'\end{tabular}                                                         \\ \hline
\begin{tabular}[c]{@{}c@{}}Topic\\ Classification\end{tabular} & \begin{tabular}[c]{@{}l@{}}Classify the following news article into one of four \\ topics: World, Sports, Business, Sci/Tech: 'The\\ central bank announced a new policy to regulate\\ inflation and stabilize the economy.'\end{tabular}                                                          \\ \hline
\begin{tabular}[c]{@{}c@{}}Text\\ Summarization\end{tabular}   & \begin{tabular}[c]{@{}l@{}}Produce text summary of the following \\ text: 'Harry Potter star Daniel Radcliffe gains access \\ to a reported Â£20 million (\$41.1 million) fortune as \\ he turns 18 on Monday, but he insists the money \\ won't cast a spell on him. ...'\end{tabular} \\ \hline
\begin{tabular}[c]{@{}c@{}}Machine\\ Translation\end{tabular}  & \begin{tabular}[c]{@{}l@{}}Translate the following sentence from English to \\ Arabic in the Qatar dialect: 'Artificial intelligence\\ is transforming the future of technology.'\end{tabular}                                                                                                     \\ \hline
\begin{tabular}[c]{@{}c@{}}Textual\\ Entailment\end{tabular}   & \begin{tabular}[c]{@{}l@{}}Given the sentences in the following format sent\_1, \\ sent\_2, determine if sent\_1 entails sent\_2 or not: \\ 'The sun is shining brightly., It is nighttime.'\end{tabular}                                                                                          \\ \hline
\end{tabular}

\end{table}

\item Manual query execution: Each query is entered manually into ChatGPT and DeepSeek to simulate real-world user interactions. The generated responses are collected and stored for further analysis.

\item Data storage and documentation: All responses are systematically documented in a structured format to facilitate the offline evaluation process. Figure \ref{smpleExc} shows a sample of responses stored for the topic classification task. 

\begin{figure}[!ht]
\centering
{\includegraphics[width=0.7\textwidth]{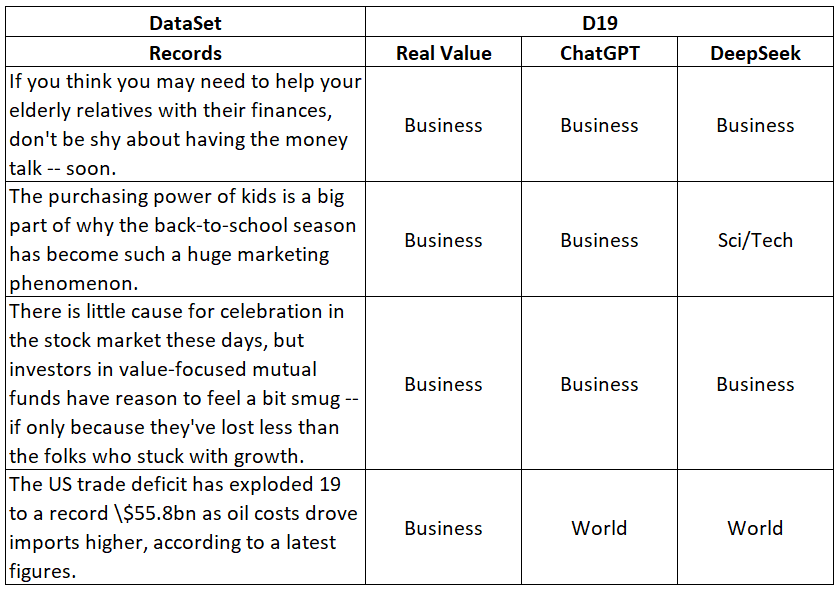}}
\caption{Sample of Collected Responses Stored.}
\label{smpleExc}
\end{figure}

\end{itemize}

\subsection{Evaluation Measures}
\label{evalM}

A set of established evaluation measures is employed to comprehensively evaluate the performance of ChatGPT and DeepSeek across five distinct NLP tasks. The selected metrics are tailored to the specific nature of each task to ensure a robust and meaningful comparison. For classification tasks, including sentiment analysis, topic classification, and textual entailment, precision, recall, F1-score, and accuracy are used to assess the models' performance. These metrics provide a balanced evaluation of how well the models identify correct instances (precision), relevant instances (recall), and their overall performance (F1-score and accuracy). For generative tasks, such as text summarization and machine translation, BERTScore is employed. BERTScore measures the quality of generated text by comparing it to reference texts, using contextual embeddings to capture semantic similarity, making it suitable for evaluating the quality and fluency of the model-generated outputs.

\subsubsection{Classification Tasks}
 
For Classification tasks, including sentiment analysis, topic classification, and textual entailment, standard classification evaluation measures are adopted:

\begin{itemize}

\item  Precision. Precision measures the proportion of correctly classified positive instances among all instances predicted as positive. A higher precision indicates fewer false positive classifications, which is critical when the cost of false positives is high. It is formally defined as:

\begin{equation}
 Precision = \frac{TP}{TP + FP}  ,
\end{equation}
  
where $TP$ represents true positives, and $FP$ denotes false positives. 

\item Recall. Recall quantifies the model’s ability to identify all relevant positive instances correctly. A higher recall means that fewer relevant instances are missed by the model, which is important when it is essential not to overlook any positive cases. It is defined as:

 \begin{equation} 
 Recall = \frac{TP}{TP+FN} ,  
 \end{equation}
 
where $FN$ represents false negatives. 

\item F1-Score. The F1-score is the harmonic mean of precision and recall, offering a balanced measure of both metrics. The F1-score is particularly useful when dealing with cases where precision and recall must be optimized simultaneously. It is computed as:

 \begin{equation}
 F1-score = \frac{2*Precision*Recall}{Precision+Recall}   ,
 \end{equation}
 
\item Accuracy. Accuracy represents the proportion of correctly classified instances in all classes. It is defined as:

\begin{equation}
 Accuracy = \frac{TP+TN}{TP+TN+FP+FN}   ,
\end{equation}
 
where $TN$ represents true negatives. 

\end{itemize}

\subsubsection{Generative Tasks}

For evaluating text generation tasks (Text Summarization and Machine Translation), the BERTScore evaluation measure is used, which provides a more semantically meaningful assessment compared to traditional n-gram-based metrics:

BERT Score evaluation measure is used to evaluate text generation tasks, including text summarization and machine translation. BERT Score is particularly effective for assessing semantic adequacy, as it captures deeper contextual meaning rather than surface-level lexical similarity. Traditional measures like BLEU and ROUGE often focus on n-gram overlap, which may fail to account for semantic variations like paraphrasing or synonym usage. In contrast, BERTScore leverages the power of pre-trained transformer models to understand the meaning behind words in context, making it highly suitable for assessing tasks like text summarization and machine translation. These tasks often involve rewording or rephrasing information, where word choice, syntactic variations, and semantic equivalence are crucial to a meaningful evaluation. BERTScore's ability to evaluate this semantic richness ensures a more accurate and relevant measure of the quality of the generated text. The measure is based on token-wise cosine similarity and is defined as:

    \begin{equation} 
     BERTScore = \frac{1}{N} \sum_{i=1}^{N} \max_{j} cosine({E({x_i})E({y_i})}) ,
    \end{equation}
    
where $E({x_i})$ and $E({y_i})$ represent contextual embeddings of tokens from the generated and reference texts, respectively.

\section{Experimental Results and Discussion}
\label{res}

This section presents the experimental results of ChatGPT and DeepSeek across five NLP tasks. The performance of LLMs is evaluated using two benchmark datasets per task. The collected responses are then analyzed to compare the effectiveness of both models.

\subsection{Sentiment Analysis Task Results}

The sentiment classification performance of ChatGPT and DeepSeek is evaluated using the IMDB and Multilingual Sentiment datasets. Table \ref{ConfSA-D3} and Table \ref{ConfSA-D2} present the confusion matrices for both models, while Table \ref{SA-D3} and Table \ref{SA-D2} report their precision, recall, F1-Score, and accuracy. Overall, DeepSeek demonstrates superior sentiment analysis performance compared to ChatGPT.

\begin{table}[h!]
\caption{Confusion Matrix of LLMs - Sentiment Analysis using IMDB dataset. (a) ChatGPT. (b) DeepSeek.}
\label{ConfSA-D3}

\begin{tabular}{ccclccc}
\multicolumn{3}{c}{(a)}                                                                        &                       & \multicolumn{3}{c}{(b)}                                                                       \\ \cline{2-3} \cline{6-7} 
\multicolumn{1}{l|}{}          & \multicolumn{1}{c|}{Positive} & \multicolumn{1}{c|}{Negative} &                       & \multicolumn{1}{l|}{}         & \multicolumn{1}{c|}{Positive} & \multicolumn{1}{c|}{Negative} \\ \cline{1-3} \cline{5-7} 
\multicolumn{1}{|c|}{Positive} & \multicolumn{1}{c|}{39}       & \multicolumn{1}{c|}{11}       & \multicolumn{1}{l|}{} & \multicolumn{1}{c|}{Positive} & \multicolumn{1}{c|}{50}       & \multicolumn{1}{c|}{0}        \\ \cline{1-3} \cline{5-7} 
\multicolumn{1}{|c|}{Negative} & \multicolumn{1}{c|}{1}        & \multicolumn{1}{c|}{49}       & \multicolumn{1}{l|}{} & \multicolumn{1}{c|}{Negative} & \multicolumn{1}{c|}{1}        & \multicolumn{1}{c|}{49}       \\ \cline{1-3} \cline{5-7} 
\end{tabular}
\end{table}

\begin{table}[h!]
\caption{Performance of LLMs - Sentiment Analysis using IMDB Dataset.}
\label{SA-D3}
\begin{tabular}{|cc|c|c|c|c|}
\hline
\multicolumn{2}{|c|}{LLM}                                  & Precision & Recall  & F1-Score & Accuracy                \\ \hline
\multicolumn{1}{|c|}{\multirow{2}{*}{ChatGPT}}  & Positive & 97.5\%    & 78.0\%  & 86.7\%   & \multirow{2}{*}{87.9\%} \\ \cline{2-5}
\multicolumn{1}{|c|}{}                          & Negative & 81.4\%    & 98.0\%  & 88.9\%   &                         \\ \hline
\multicolumn{1}{|c|}{\multirow{2}{*}{DeepSeek}} & Positive & 98.0\%    & 100.0\% & 99.0\%   & \multirow{2}{*}{99.0\%} \\ \cline{2-5}
\multicolumn{1}{|c|}{}                          & Negative & 100.0\%   & 98.0\%  & 99.0\%   &                         \\ \hline
\end{tabular}
\end{table}

Table \ref{ConfSA-D3} and Table \ref{SA-D3} present the results collected from both LLMs using the IMDB dataset. The sentiment class in the IMDB dataset is a binary, where the review is either positive or negative. Notably, DeepSeek achieves a higher overall accuracy (99.0\%) compared to ChatGPT (87.9\%). The recall for positive sentiment is 100\% for DeepSeek, indicating that it classified all positive samples correctly, whereas ChatGPT achieves 78.0\%. Similarly, DeepSeek attains higher F1-Scores across both sentiment classes, reinforcing its superior performance in binary sentiment classification. These results prove that DeepSeek outperforms ChatGPT in sentiment analysis for this dataset, particularly in detecting positive sentiment, which is crucial in applications such as customer feedback analysis and opinion mining.

\begin{table}[h!]
\caption{Confusion Matrix of LLMs - Sentiment Analysis using Multilingual Sentiment Dataset. (a) ChatGPT. (b) DeepSeek.}
\label{ConfSA-D2}
\begin{tabular}{cccc}
\multicolumn{4}{c}{(a)}                                                                                                       \\ \cline{2-4} 
\multicolumn{1}{l|}{}          & \multicolumn{1}{c|}{Positive} & \multicolumn{1}{c|}{Neutral} & \multicolumn{1}{c|}{Negative} \\ \hline
\multicolumn{1}{|c|}{Positive} & \multicolumn{1}{c|}{34}       & \multicolumn{1}{c|}{13}      & \multicolumn{1}{c|}{3}        \\ \hline
\multicolumn{1}{|c|}{Neutral}  & \multicolumn{1}{c|}{4}        & \multicolumn{1}{c|}{28}      & \multicolumn{1}{c|}{18}       \\ \hline
\multicolumn{1}{|c|}{Negative} & \multicolumn{1}{c|}{0}        & \multicolumn{1}{c|}{15}      & \multicolumn{1}{c|}{35}       \\ \hline
\multicolumn{1}{l}{}           & \multicolumn{1}{l}{}          & \multicolumn{1}{l}{}         & \multicolumn{1}{l}{}          \\
\multicolumn{4}{c}{(b)}                                                                                                       \\ \cline{2-4} 
\multicolumn{1}{l|}{}          & \multicolumn{1}{c|}{Positive} & \multicolumn{1}{c|}{Neutral} & \multicolumn{1}{c|}{Negative} \\ \hline
\multicolumn{1}{|c|}{Positive} & \multicolumn{1}{c|}{44}       & \multicolumn{1}{c|}{6}       & \multicolumn{1}{c|}{0}        \\ \hline
\multicolumn{1}{|c|}{Neutral}  & \multicolumn{1}{c|}{5}        & \multicolumn{1}{c|}{36}      & \multicolumn{1}{c|}{9}        \\ \hline
\multicolumn{1}{|c|}{Negative} & \multicolumn{1}{c|}{2}        & \multicolumn{1}{c|}{14}      & \multicolumn{1}{c|}{34}       \\ \hline
\end{tabular}
\end{table}

\begin{table}[h!]
\caption{Performance of LLMs - Sentiment Analysis using Multilingual Sentiment Dataset.}
\label{SA-D2}
\begin{tabular}{|cc|c|c|c|c|}
\hline
\multicolumn{2}{|c|}{LLM}                                  & Precision & Recall & F1-Score & Accuracy                \\ \hline
\multicolumn{1}{|c|}{\multirow{3}{*}{ChatGPT}}  & Positive & 85.0\%    & 68.0\% & 75.6\%   & \multirow{3}{*}{64.7\%} \\ \cline{2-5}
\multicolumn{1}{|c|}{}                          & Neutral  & 50.0\%    & 56.0\% & 52.8\%   &                         \\ \cline{2-5}
\multicolumn{1}{|c|}{}                          & Negative & 62.5\%    & 70.0\% & 66.0\%   &                         \\ \hline
\multicolumn{1}{|c|}{\multirow{3}{*}{DeepSeek}} & Positive & 86.3\%    & 88.0\% & 87.1\%   & \multirow{3}{*}{76.0\%} \\ \cline{2-5}
\multicolumn{1}{|c|}{}                          & Neutral  & 64.3\%    & 72.0\% & 67.9\%   &                         \\ \cline{2-5}
\multicolumn{1}{|c|}{}                          & Negative & 79.1\%    & 68.0\% & 73.1\%   &                         \\ \hline
\end{tabular}
\end{table}

Table \ref{ConfSA-D2} and Table \ref{SA-D2} present the results collected from both LLMs using the Multilingual Sentiment dataset. This dataset is used to evaluate the LLMs' ability to handle more complex sentiment classification, which introduces an additional "neutral" sentiment class. The confusion matrices in Table \ref{ConfSA-D2} highlight key performance differences between ChatGPT and DeepSeek. ChatGPT shows a modest performance with neutral sentiment, misclassifying a significant number of neutral samples as either positive or negative (eighteen as negative and four as positive). On the other hand, DeepSeek demonstrates a more stable classification performance, showing fewer misclassifications in the neutral category, contributing to its higher overall accuracy. While DeepSeek achieves an accuracy of 76.0\%, ChatGPT lags at 64.7\%. ChatGPT's F1-Score for neutral sentiment is significantly lower (52.8\%) compared to DeepSeek (67.9\%), highlighting its difficulty in handling this additional sentiment category. Furthermore, DeepSeek achieves consistently higher precision, recall, and F-score values across all sentiment classes.

\subsection{Topic Classification Task Results}

ChatGPT and DeepSeek are assessed in topic classification using the AG News Classification and Web of Science datasets. Table \ref{ConfTC-D19} and Table \ref{ConfTC-D20} show the confusion matrices for both LLMs. Table \ref{TC-D19} and Table \ref{TC-D20} summarize their precision, recall, F1-score, and accuracy. The results indicate that DeepSeek achieves higher performance than ChatGPT in topic classification.

\begin{table}[h!]
\caption{Confusion Matrix of LLMs - Topic Classification using AG News Classification Dataset. (a) ChatGPT. (b) DeepSeek.}
\label{ConfTC-D19}
\begin{tabular}{ccccc}
\multicolumn{5}{c}{(a)}                                                                                                                                   \\ \cline{2-5} 
\multicolumn{1}{l|}{}          & \multicolumn{1}{c|}{Business} & \multicolumn{1}{c|}{Sci/Tech} & \multicolumn{1}{c|}{Sport} & \multicolumn{1}{c|}{World} \\ \hline
\multicolumn{1}{|c|}{Business} & \multicolumn{1}{c|}{11}       & \multicolumn{1}{c|}{0}        & \multicolumn{1}{c|}{0}      & \multicolumn{1}{c|}{9}     \\ \hline
\multicolumn{1}{|c|}{Sci/Tech} & \multicolumn{1}{c|}{0}        & \multicolumn{1}{c|}{17}       & \multicolumn{1}{c|}{3}      & \multicolumn{1}{c|}{0}     \\ \hline
\multicolumn{1}{|c|}{Sport}   & \multicolumn{1}{c|}{0}        & \multicolumn{1}{c|}{0}        & \multicolumn{1}{c|}{18}     & \multicolumn{1}{c|}{2}     \\ \hline
\multicolumn{1}{|c|}{World}    & \multicolumn{1}{c|}{0}        & \multicolumn{1}{c|}{0}        & \multicolumn{1}{c|}{2}      & \multicolumn{1}{c|}{18}    \\ \hline
\multicolumn{1}{l}{}           & \multicolumn{1}{l}{}          & \multicolumn{1}{l}{}          & \multicolumn{1}{l}{}        & \multicolumn{1}{l}{}       \\
\multicolumn{5}{c}{(b)}                                                                                                                                   \\ \cline{2-5} 
\multicolumn{1}{l|}{}          & \multicolumn{1}{c|}{Business} & \multicolumn{1}{c|}{Sci/Tech} & \multicolumn{1}{c|}{Sport} & \multicolumn{1}{c|}{World} \\ \hline
\multicolumn{1}{|c|}{Business} & \multicolumn{1}{c|}{13}       & \multicolumn{1}{c|}{2}        & \multicolumn{1}{c|}{0}      & \multicolumn{1}{c|}{5}     \\ \hline
\multicolumn{1}{|c|}{Sci/Tech} & \multicolumn{1}{c|}{3}        & \multicolumn{1}{c|}{14}       & \multicolumn{1}{c|}{3}      & \multicolumn{1}{c|}{0}     \\ \hline
\multicolumn{1}{|c|}{Sport}   & \multicolumn{1}{c|}{0}        & \multicolumn{1}{c|}{0}        & \multicolumn{1}{c|}{20}     & \multicolumn{1}{c|}{0}     \\ \hline
\multicolumn{1}{|c|}{World}    & \multicolumn{1}{c|}{0}        & \multicolumn{1}{c|}{0}        & \multicolumn{1}{c|}{2}      & \multicolumn{1}{c|}{18}    \\ \hline
\end{tabular}
\end{table}

\begin{table}[h!]
\caption{Performance of LLMs - Topic Classification using AG News Classification Dataset.}
\label{TC-D19}
\begin{tabular}{|cc|c|c|c|c|}
\hline
\multicolumn{2}{|c|}{LLM}                                  & Precision & Recall  & F1-Score & Accuracy                \\ \hline
\multicolumn{1}{|c|}{\multirow{4}{*}{ChatGPT}}  & Business & 100.0\%   & 55.0\%  & 71.0\%   & \multirow{4}{*}{80.0\%} \\ \cline{2-5}
\multicolumn{1}{|c|}{}                          & Sci/Tech & 100.0\%   & 85.0\%  & 91.9\%   &                         \\ \cline{2-5}
\multicolumn{1}{|c|}{}                          & Sport    & 78.3\%    & 90.0\%  & 83.7\%   &                         \\ \cline{2-5}
\multicolumn{1}{|c|}{}                          & World    & 62.1\%    & 90.0\%  & 73.5\%   &                         \\ \hline
\multicolumn{1}{|c|}{\multirow{4}{*}{DeepSeek}} & Business & 81.3\%    & 65.0\%  & 72.2\%   & \multirow{4}{*}{81.3\%} \\ \cline{2-5}
\multicolumn{1}{|c|}{}                          & Sci/Tech & 87.5\%    & 70.0\%  & 77.8\%   &                         \\ \cline{2-5}
\multicolumn{1}{|c|}{}                          & Sport    & 80.0\%    & 100.0\% & 88.9\%   &                         \\ \cline{2-5}
\multicolumn{1}{|c|}{}                          & World    & 78.3\%    & 90.0\%  & 83.7\%   &                         \\ \hline
\end{tabular}
\end{table}

Table \ref{ConfTC-D19} and Table \ref{TC-D19} present the results collected from ChatGPT and DeepSeek using the AG News Classification dataset. This dataset comprises four categories: Business, Sci/Tech, Sport, and World. The confusion matrices in Table \ref{ConfSA-D2} highlight key classification patterns and challenges for each model. ChatGPT exhibits better performance in classifying Sci/Tech news articles than DeepSeek. In contrast, DeepSeek shows fewer misclassifications in the Business and Sport classes compared to ChatGPT. DeepSeek achieved a higher accuracy (81.3\%) than ChatGPT (80\%). ChatGPT shows high precision for Business and Sci/Tech (100\%), while it exhibits a low precision for World class (62.1\%). On average, ChatGPT's F1-Score is lower (80.03\%) compared to DeepSeek (80.65\%). ChatGPT generally had better precision, making it more reliable when misclassifications need to be minimized. On the other hand, DeepSeek had a better recall, indicating it retrieved more correct instances per category.

\begin{table}[h!]
\caption{Confusion Matrix of LLMs - Topic Classification using Web of Science  Dataset. (a) ChatGPT. (b) DeepSeek.}
\label{ConfTC-D20}
\resizebox{\columnwidth}{!}{
\begin{tabular}{cccccccc}
\multicolumn{8}{c}{(a)}                                                                                                                                                                                                                              \\ \cline{2-8} 
\multicolumn{1}{l|}{}              & \multicolumn{1}{c|}{Biochemistry} & \multicolumn{1}{c|}{Civil} & \multicolumn{1}{c|}{CS} & \multicolumn{1}{c|}{ECE} & \multicolumn{1}{c|}{MAE} & \multicolumn{1}{c|}{Medical} & \multicolumn{1}{c|}{Psychology} \\ \hline
\multicolumn{1}{|c|}{Biochemistry} & \multicolumn{1}{c|}{25}           & \multicolumn{1}{c|}{0}     & \multicolumn{1}{c|}{0}  & \multicolumn{1}{c|}{4}   & \multicolumn{1}{c|}{0}   & \multicolumn{1}{c|}{7}       & \multicolumn{1}{c|}{1}          \\ \hline
\multicolumn{1}{|c|}{Civil}        & \multicolumn{1}{c|}{1}            & \multicolumn{1}{c|}{2}     & \multicolumn{1}{c|}{1}  & \multicolumn{1}{c|}{1}   & \multicolumn{1}{c|}{2}   & \multicolumn{1}{c|}{0}       & \multicolumn{1}{c|}{0}          \\ \hline
\multicolumn{1}{|c|}{CS}           & \multicolumn{1}{c|}{0}            & \multicolumn{1}{c|}{0}     & \multicolumn{1}{c|}{15} & \multicolumn{1}{c|}{1}   & \multicolumn{1}{c|}{1}   & \multicolumn{1}{c|}{1}       & \multicolumn{1}{c|}{0}          \\ \hline
\multicolumn{1}{|c|}{ECE}          & \multicolumn{1}{c|}{0}            & \multicolumn{1}{c|}{1}     & \multicolumn{1}{c|}{0}  & \multicolumn{1}{c|}{3}   & \multicolumn{1}{c|}{1}   & \multicolumn{1}{c|}{0}       & \multicolumn{1}{c|}{1}          \\ \hline
\multicolumn{1}{|c|}{MAE}          & \multicolumn{1}{c|}{2}            & \multicolumn{1}{c|}{1}     & \multicolumn{1}{c|}{0}  & \multicolumn{1}{c|}{1}   & \multicolumn{1}{c|}{1}   & \multicolumn{1}{c|}{0}       & \multicolumn{1}{c|}{0}          \\ \hline
\multicolumn{1}{|c|}{Medical}      & \multicolumn{1}{c|}{11}           & \multicolumn{1}{c|}{0}     & \multicolumn{1}{c|}{0}  & \multicolumn{1}{c|}{1}   & \multicolumn{1}{c|}{0}   & \multicolumn{1}{c|}{9}       & \multicolumn{1}{c|}{3}          \\ \hline
\multicolumn{1}{|c|}{Psychology}   & \multicolumn{1}{c|}{1}            & \multicolumn{1}{c|}{0}     & \multicolumn{1}{c|}{0}  & \multicolumn{1}{c|}{0}   & \multicolumn{1}{c|}{0}   & \multicolumn{1}{c|}{1}       & \multicolumn{1}{c|}{1}          \\ \hline
\multicolumn{1}{l}{}               & \multicolumn{1}{l}{}              & \multicolumn{1}{l}{}       & \multicolumn{1}{l}{}    & \multicolumn{1}{l}{}     & \multicolumn{1}{l}{}     & \multicolumn{1}{l}{}         & \multicolumn{1}{l}{}            \\
\multicolumn{8}{c}{(b)}                                                                                                                                                                                                                              \\ \cline{2-8} 
\multicolumn{1}{l|}{}              & \multicolumn{1}{c|}{Biochemistry} & \multicolumn{1}{c|}{Civil} & \multicolumn{1}{c|}{CS} & \multicolumn{1}{c|}{ECE} & \multicolumn{1}{c|}{MAE} & \multicolumn{1}{c|}{Medical} & \multicolumn{1}{c|}{Psychology} \\ \hline
\multicolumn{1}{|c|}{Biochemistry} & \multicolumn{1}{c|}{19}           & \multicolumn{1}{c|}{1}     & \multicolumn{1}{c|}{1}  & \multicolumn{1}{c|}{1}   & \multicolumn{1}{c|}{0}   & \multicolumn{1}{c|}{13}      & \multicolumn{1}{c|}{2}          \\ \hline
\multicolumn{1}{|c|}{Civil}        & \multicolumn{1}{c|}{0}            & \multicolumn{1}{c|}{4}     & \multicolumn{1}{c|}{1}  & \multicolumn{1}{c|}{2}   & \multicolumn{1}{c|}{0}   & \multicolumn{1}{c|}{0}       & \multicolumn{1}{c|}{0}          \\ \hline
\multicolumn{1}{|c|}{CS}           & \multicolumn{1}{c|}{1}            & \multicolumn{1}{c|}{1}     & \multicolumn{1}{c|}{12} & \multicolumn{1}{c|}{2}   & \multicolumn{1}{c|}{0}   & \multicolumn{1}{c|}{2}       & \multicolumn{1}{c|}{0}          \\ \hline
\multicolumn{1}{|c|}{ECE}          & \multicolumn{1}{c|}{0}            & \multicolumn{1}{c|}{0}     & \multicolumn{1}{c|}{1}  & \multicolumn{1}{c|}{2}   & \multicolumn{1}{c|}{1}   & \multicolumn{1}{c|}{2}       & \multicolumn{1}{c|}{0}          \\ \hline
\multicolumn{1}{|c|}{MAE}          & \multicolumn{1}{c|}{2}            & \multicolumn{1}{c|}{3}     & \multicolumn{1}{c|}{0}  & \multicolumn{1}{c|}{0}   & \multicolumn{1}{c|}{0}   & \multicolumn{1}{c|}{0}       & \multicolumn{1}{c|}{0}          \\ \hline
\multicolumn{1}{|c|}{Medical}      & \multicolumn{1}{c|}{7}            & \multicolumn{1}{c|}{0}     & \multicolumn{1}{c|}{0}  & \multicolumn{1}{c|}{0}   & \multicolumn{1}{c|}{0}   & \multicolumn{1}{c|}{15}      & \multicolumn{1}{c|}{2}          \\ \hline
\multicolumn{1}{|c|}{Psychology}   & \multicolumn{1}{c|}{1}            & \multicolumn{1}{c|}{0}     & \multicolumn{1}{c|}{0}  & \multicolumn{1}{c|}{0}   & \multicolumn{1}{c|}{0}   & \multicolumn{1}{c|}{2}       & \multicolumn{1}{c|}{0}          \\ \hline
\end{tabular}
}
\end{table}

\begin{table}[h!]
\caption{Performance of LLMs - Topic Classification using Web of Science Dataset.}
\label{TC-D20}
\resizebox{\columnwidth}{!}{
\begin{tabular}{|cc|c|c|c|c|}
\hline
\multicolumn{2}{|c|}{LLM}                                      & Precision & Recall & F1-Score & Accuracy                \\ \hline
\multicolumn{1}{|c|}{\multirow{7}{*}{ChatGPT}}  & Biochemistry & 62.5\%    & 67.6\% & 64.9\%   & \multirow{7}{*}{56.0\%} \\ \cline{2-5}
\multicolumn{1}{|c|}{}                          & Civil        & 50.0\%    & 28.6\% & 36.4\%   &                         \\ \cline{2-5}
\multicolumn{1}{|c|}{}                          & CS           & 93.8\%    & 83.3\% & 88.2\%   &                         \\ \cline{2-5}
\multicolumn{1}{|c|}{}                          & ECE          & 27.3\%    & 50.0\% & 35.3\%   &                         \\ \cline{2-5}
\multicolumn{1}{|c|}{}                          & MAE          & 20.0\%    & 20.0\% & 20.0\%   &                         \\ \cline{2-5}
\multicolumn{1}{|c|}{}                          & Medical      & 50.0\%    & 37.5\% & 42.9\%   &                         \\ \cline{2-5}
\multicolumn{1}{|c|}{}                          & Psychology   & 16.7\%    & 33.3\% & 22.2\%   &                         \\ \hline
\multicolumn{1}{|c|}{\multirow{7}{*}{DeepSeek}} & Biochemistry & 63.3\%    & 51.4\% & 56.7\%   & \multirow{7}{*}{52.0\%} \\ \cline{2-5}
\multicolumn{1}{|c|}{}                          & Civil        & 44.4\%    & 57.1\% & 50.0\%   &                         \\ \cline{2-5}
\multicolumn{1}{|c|}{}                          & CS           & 80.0\%    & 66.7\% & 72.7\%   &                         \\ \cline{2-5}
\multicolumn{1}{|c|}{}                          & ECE          & 28.6\%    & 33.3\% & 30.8\%   &                         \\ \cline{2-5}
\multicolumn{1}{|c|}{}                          & MAE          & 0.0\%     & 0.0\%  & 0.0\%    &                         \\ \cline{2-5}
\multicolumn{1}{|c|}{}                          & Medical      & 44.1\%    & 62.5\% & 51.7\%   &                         \\ \cline{2-5}
\multicolumn{1}{|c|}{}                          & Psychology   & 0.0\%     & 0.0\%  & 0.0\%    &                         \\ \hline
\end{tabular}
}
\end{table}

Table \ref{ConfTC-D20}  and Table \ref{TC-D20} present the results of topic classification using the Web of Science dataset for both ChatGPT and DeepSeek. This dataset includes seven diverse scientific disciplines, posing a challenge for general-purpose LLMs. Table \ref{ConfTC-D20} highlights key classification patterns and challenges for each model. ChatGPT performs better in classifying CS than DeepSeek, correctly identifying 15 out of 18 instances. Similarly, Biochemistry is identified with moderate accuracy (25 correct predictions out of 37). On the other hand, ChatGPT demonstrates a higher misclassification rate in the Civil class compared to DeepSeek. A critical observation is DeepSeek’s complete failure to classify Psychology and MAE, as all samples from these categories are misclassified into others, leading to an F1-score of 0.0\%. Furthermore, Medical samples are predominantly classified correctly (15 out of 18), contributing to higher recall in this category than ChatGPT. Overall, ChatGPT achieves a higher overall accuracy (56.0\%) than DeepSeek (52.0\%). For CS and Biochemistry, ChatGPT demonstrates a better F1-Score (88.2\% and 64.9\%, respectively) than DeepSeek (72.7\% and 56.7\%, respectively). However, ChatGPT faces challenges in classifying Civil (recall of 28.6\%) and Medical (recall of 37.5\%) scientific articles. In contrast, DeepSeek exhibits better recall in Medical (62.5\%) and Civil (57.1\%). Both LLMs generally struggle with niche fields like MAE and Psychology, suggesting domain-specific adaptation is needed for improved classification.

\subsection{Text Summarization Task Results}

ChatGPT and DeepSeek are assessed in text summarization task using Gigaword and CNN/Daily Mail datasets. The evaluation was carried out using the BERT Score, which includes three measures: precision, recall, and F1-Score. Table \ref{TS-D14} and Table \ref{TS-D13} summarize the results of these evaluations.

\begin{table}[h!]
\caption{Performance of LLMs - Text Summarization using Gigaword Dataset.}
\label{TS-D14}
\begin{tabular}{|c|ccc|}
\hline
\multirow{2}{*}{LLM} & \multicolumn{3}{c|}{BERT Score}                                          \\ \cline{2-4} 
                     & \multicolumn{1}{c|}{Precision} & \multicolumn{1}{c|}{Recall}  & F1-Score \\ \hline
ChatGPT              & \multicolumn{1}{c|}{71.00\%}   & \multicolumn{1}{c|}{72.28\%} & 71.59\%  \\ \hline
DeepSeek             & \multicolumn{1}{c|}{70.62\%}   & \multicolumn{1}{c|}{71.66\%} & 71.11\%  \\ \hline
\end{tabular}
\end{table}

Table \ref{TS-D14} presents the performance results of ChatGPT and DeepSeek for the Gigaword dataset. The Gigaword dataset consists of news articles and corresponding summaries, challenging both models to generate concise yet accurate summaries. ChatGPT slightly outperformed DeepSeek in all three BERT Score metrics. Specifically, ChatGPT achieved a precision of 71.00\%, a recall of 72.28\%, and an F1-Score of 71.59\%. In comparison, DeepSeek obtained a precision of 70.62\%, a recall of 71.66\%, and an F1-Score of 71.11\%. The results indicate that ChatGPT performs marginally better in precision and F1-Score. Therefore, ChatGPT better captures relevant content from the original text and includes it in the summarized output. Although the differences are minor, the slight edge in performance for ChatGPT could be attributed to its ability to balance content inclusion (recall) and conciseness (precision) more effectively.

\begin{table}[h!]
\caption{Performance of LLMs - Text Summarization using CNN/Daily Mail Summarization Dataset.}
\label{TS-D13}
\begin{tabular}{|c|ccc|}
\hline
\multirow{2}{*}{LLM} & \multicolumn{3}{c|}{BERT Score}                                          \\ \cline{2-4} 
                     & \multicolumn{1}{c|}{Precision} & \multicolumn{1}{c|}{Recall}  & F1-Score \\ \hline
ChatGPT              & \multicolumn{1}{c|}{68.80\%}   & \multicolumn{1}{c|}{73.59\%} & 71.09\%  \\ \hline
DeepSeek             & \multicolumn{1}{c|}{68.44\%}   & \multicolumn{1}{c|}{74.11\%} & 71.13\%  \\ \hline
\end{tabular}
\end{table}

Table \ref{TS-D13} presents the performance results of ChatGPT and DeepSeek for the CNN/Daily Mail dataset. The CNN/Daily Mail dataset is a widely used text summarization benchmark involving news articles and summaries requiring high levels of abstraction and content condensation. DeepSeek marginally outperformed ChatGPT in F1-Score. DeepSeek achieves a precision of 68.44\%, a recall of 74.11\%, and an F1-Score of 71.13\%. In contrast, ChatGPT shows a precision of 68.80\%, a recall of 73.59\%, and an F1-Score of 71.09\%. While ChatGPT demonstrates slightly higher precision, DeepSeek exhibits a marginally better recall and F1-Score. These results indicate that DeepSeek may be more effective at generating summaries with broader content coverage. On the other hand, ChatGPT excels in producing more concise summaries and performing better in terms of output brevity.

\subsection{Machine Translation Task Results}

The performance of ChatGPT and DeepSeek is evaluated on the Machine Translation task, which involved translating from English to Arabic. Two datasets are used for this evaluation: ArzEn-MultiGenre and AraBench. The LLMs are assessed using the BERT Score, which includes three measures: precision, recall, and F1-Score. Table \ref{MT-D7} and Table \ref{MT-D6} summarize the results for the respective datasets.

\begin{table}[h!]
\caption{Performance of LLMs - Machine Translation using ArzEn-MultiGenre Dataset.}
\label{MT-D7}
\begin{tabular}{|c|ccc|}
\hline
\multirow{2}{*}{LLM} & \multicolumn{3}{c|}{BERT Score}                                          \\ \cline{2-4} 
                     & \multicolumn{1}{c|}{Precision} & \multicolumn{1}{c|}{Recall}  & F1-Score \\ \hline
ChatGPT              & \multicolumn{1}{c|}{78.23\%}   & \multicolumn{1}{c|}{78.61\%} & 78.39\%  \\ \hline
DeepSeek             & \multicolumn{1}{c|}{77.80\%}   & \multicolumn{1}{c|}{77.32\%} & 77.53\%  \\ \hline
\end{tabular}
\end{table}

Table \ref{MT-D7} presents the performance results of ChatGPT and DeepSeek for the ArzEn-MultiGenre dataset. This dataset consists of Egyptian Arabic song lyrics, novels, and subtitles paired with English translations. ChatGPT outperformed DeepSeek in all three BERT Score metrics. Specifically, ChatGPT achieved a precision of 78.23\%, recall of 78.61\%, and an F1-Score of 78.39\%. In comparison, DeepSeek attained a precision of 77.80\%, recall of 77.32\%, and an F1-Score of 77.53\%. The results indicate that ChatGPT performs slightly better than DeepSeek across all BERT Score metrics. Hence, ChatGPT may more effectively capture relevant translation content in the generated text, leading to a higher recall value.

\begin{table}[h!]
\caption{Performance of LLMs - Machine Translation using AraBench Dataset.}
\label{MT-D6}
\begin{tabular}{|c|ccc|}
\hline
\multirow{2}{*}{LLM} & \multicolumn{3}{c|}{BERT Score}                                          \\ \cline{2-4} 
                     & \multicolumn{1}{c|}{Precision} & \multicolumn{1}{c|}{Recall}  & F1-Score \\ \hline
ChatGPT              & \multicolumn{1}{c|}{78.34\%}   & \multicolumn{1}{c|}{78.37\%} & 78.33\%  \\ \hline
DeepSeek             & \multicolumn{1}{c|}{78.79\%}   & \multicolumn{1}{c|}{78.60\%} & 78.67\%  \\ \hline
\end{tabular}
\end{table}

Table \ref{MT-D6} presents the performance results of ChatGPT and DeepSeek for the AraBench dataset, which includes Arabic-English translation pairs across various dialects. The evaluation focuses on Qatari and Jordanian Arabic. DeepSeek showed slight superiority in all three BERT Score metrics, achieving a precision of 78.79\%, a recall of 78.60\%, and an F1-Score of 78.67\%. Conversely, ChatGPT achieved a precision of 78.34\%, a recall of 78.37\%, and an F1-Score of 78.33\%. The results indicate that both LLMs perform similarly well in translating between English and Arabic, mainly when dealing with dialects such as Qatari and Jordanian Arabic.

\subsection{Textual Entailment Task Results}

The textual entailment performance of ChatGPT and DeepSeek is evaluated using the Scitail and FraCaS datasets. Table \ref{ConfTE-D17} and Table \ref{ConfTE-D18} present the confusion matrices for both LLMs. Table \ref{TE-D17} and Table \ref{TE-D18} report their precision, recall, F1-Score, and accuracy. Overall, DeepSeek outperforms ChatGPT in textual entailment tasks.

\begin{table}[h!]
\caption{Confusion Matrix of LLMs - Textual Entailment using Scitail Dataset. (a) ChatGPT. (b) DeepSeek.}
\label{ConfTE-D17}
\begin{tabular}{ccclccc}
\multicolumn{3}{c}{(a)}                                                                    &                       & \multicolumn{3}{c}{(b)}                                                                   \\ \cline{2-3} \cline{6-7} 
\multicolumn{1}{l|}{}         & \multicolumn{1}{c|}{Entail} & \multicolumn{1}{c|}{Neutral} &                       & \multicolumn{1}{l|}{}        & \multicolumn{1}{c|}{Entail} & \multicolumn{1}{c|}{Neutral} \\ \cline{1-3} \cline{5-7} 
\multicolumn{1}{|c|}{Entail}  & \multicolumn{1}{c|}{25}     & \multicolumn{1}{c|}{0}       & \multicolumn{1}{l|}{} & \multicolumn{1}{c|}{Entail}  & \multicolumn{1}{c|}{25}     & \multicolumn{1}{c|}{0}       \\ \cline{1-3} \cline{5-7} 
\multicolumn{1}{|c|}{Neutral} & \multicolumn{1}{c|}{20}     & \multicolumn{1}{c|}{5}       & \multicolumn{1}{l|}{} & \multicolumn{1}{c|}{Neutral} & \multicolumn{1}{c|}{18}     & \multicolumn{1}{c|}{7}       \\ \cline{1-3} \cline{5-7} 
\end{tabular}
\end{table}

\begin{table}[h!]
\caption{Performance of LLMs - Textual Entailment using Scitail Dataset.}
\label{TE-D17}
\begin{tabular}{|cc|c|c|c|c|}
\hline
\multicolumn{2}{|c|}{LLM}                                 & Precision & Recall  & F1-Score & Accuracy                \\ \hline
\multicolumn{1}{|c|}{\multirow{2}{*}{ChatGPT}}  & Entail  & 55.6\%    & 100.0\% & 71.4\%   & \multirow{2}{*}{60.0\%} \\ \cline{2-5}
\multicolumn{1}{|c|}{}                          & Neutral & 100.0\%   & 20.0\%  & 33.3\%   &                         \\ \hline
\multicolumn{1}{|c|}{\multirow{2}{*}{DeepSeek}} & Entail  & 58.1\%    & 100.0\% & 73.5\%   & \multirow{2}{*}{64.0\%} \\ \cline{2-5}
\multicolumn{1}{|c|}{}                          & Neutral & 100.0\%   & 28.0\%  & 43.8\%   &                         \\ \hline
\end{tabular}
\end{table}

Table \ref{ConfTE-D17} and Table \ref{TE-D17} present the results collected from both LLMs using the Scitail dataset. The entailment class in the Scitail dataset is binary, where each text pair is classified as entailment or neutral. The confusion matrices in Table \ref{ConfTE-D17} highlight key performance differences between ChatGPT and DeepSeek. ChatGPT and DeepSeek correctly identify 25 instances as entailments (True Positives), with no misclassifications as neutral. For the neutral class, ChatGPT makes fewer misclassifications (5 out of 25) compared to DeepSeek. DeepSeek achieves an accuracy of 64.0\%, while ChatGPT's accuracy is 60.0\%. In terms of the entail class, DeepSeek outperforms ChatGPT with a higher precision (58.1\%) and F1-Score (73.5\%) compared to ChatGPT's precision of 55.6\% and F1-Score of 71.4\%. For the neutral class, ChatGPT has a lower recall (20.0\%) and F1-Score (33.3\%) compared to DeepSeek, which achieves a recall of 28.0\% and F1-Score of 43.8\%.

\begin{table}[h!]
\caption{Confusion Matrix of LLMs - Textual Entailment using FraCaS Dataset. (a) ChatGPT. (b) DeepSeek.}
\label{ConfTE-D18}
\begin{tabular}{cccc}
\multicolumn{4}{c}{(a)}                                                                                                         \\ \cline{2-4} 
\multicolumn{1}{l|}{}            & \multicolumn{1}{c|}{Entail} & \multicolumn{1}{c|}{Neutral} & \multicolumn{1}{c|}{Contradict} \\ \hline
\multicolumn{1}{|c|}{Entail}     & \multicolumn{1}{c|}{16}     & \multicolumn{1}{c|}{5}       & \multicolumn{1}{c|}{4}          \\ \hline
\multicolumn{1}{|c|}{Neutral}    & \multicolumn{1}{c|}{10}     & \multicolumn{1}{c|}{13}      & \multicolumn{1}{c|}{2}          \\ \hline
\multicolumn{1}{|c|}{Contradict} & \multicolumn{1}{c|}{1}      & \multicolumn{1}{c|}{5}       & \multicolumn{1}{c|}{19}         \\ \hline
\multicolumn{1}{l}{}             & \multicolumn{1}{l}{}        & \multicolumn{1}{l}{}         & \multicolumn{1}{l}{}            \\
\multicolumn{4}{c}{(b)}                                                                                                         \\ \cline{2-4} 
\multicolumn{1}{l|}{}            & \multicolumn{1}{c|}{Entail} & \multicolumn{1}{c|}{Neutral} & \multicolumn{1}{c|}{Contradict} \\ \hline
\multicolumn{1}{|c|}{Entail}     & \multicolumn{1}{c|}{20}     & \multicolumn{1}{c|}{4}       & \multicolumn{1}{c|}{1}          \\ \hline
\multicolumn{1}{|c|}{Neutral}    & \multicolumn{1}{c|}{11}     & \multicolumn{1}{c|}{10}      & \multicolumn{1}{c|}{4}          \\ \hline
\multicolumn{1}{|c|}{Contradict} & \multicolumn{1}{c|}{2}      & \multicolumn{1}{c|}{1}       & \multicolumn{1}{c|}{22}         \\ \hline
\end{tabular}
\end{table}

\begin{table}[h!]
\caption{Performance of LLMs - Textual Entailment using FraCaS Dataset.}
\label{TE-D18}
\begin{tabular}{|cc|c|c|c|c|}
\hline
\multicolumn{2}{|c|}{LLM}                                    & Precision & Recall & F1-Score & Accuracy                \\ \hline
\multicolumn{1}{|c|}{\multirow{3}{*}{ChatGPT}}  & Entail     & 59.3\%    & 64.0\% & 61.5\%   & \multirow{3}{*}{64.0\%} \\ \cline{2-5}
\multicolumn{1}{|c|}{}                          & Neutral    & 56.5\%    & 52.0\% & 54.2\%   &                         \\ \cline{2-5}
\multicolumn{1}{|c|}{}                          & Contradict & 76.0\%    & 76.0\% & 76.0\%   &                         \\ \hline
\multicolumn{1}{|c|}{\multirow{3}{*}{DeepSeek}} & Entail     & 60.6\%    & 80.0\% & 69.0\%   & \multirow{3}{*}{69.3\%} \\ \cline{2-5}
\multicolumn{1}{|c|}{}                          & Neutral    & 66.7\%    & 40.0\% & 50.0\%   &                         \\ \cline{2-5}
\multicolumn{1}{|c|}{}                          & Contradict & 81.5\%    & 88.0\% & 84.6\%   &                         \\ \hline
\end{tabular}
\end{table}

Table \ref{ConfTE-D18} and Table \ref{TE-D18} present the results collected from ChatGPT and DeepSeek using the FraCaS dataset. This dataset evaluates the LLMs' ability to handle more complex entailment classification, introducing an additional "Contradicts" class. Table \ref{ConfTE-D18} presents the confusion matrices for both models on the FraCaS dataset. DeepSeek correctly identifies 20 out of 25 instances as entailments, while ChatGPT correctly identifies 16. Similarly, DeepSeek correctly specifies 22 out of 25 contradictions, while ChatGPT correctly identifies 19. ChatGPT misclassifies 12 instances for the neural class and correctly identifies 13 instances. DeepSeek misclassifies 14 entailments as neutral and correctly identifies 10 neutral relations. The overall accuracy of DeepSeek is 69.3\%, outperforming ChatGPT (64.0\%). For entailment, ChatGPT achieves a precision of 59.3\% and a recall of 64.0\%. ChatGPT performs less effectively for the neutral class, with a precision of 56.5\% and a recall of 52.0\%. However, ChatGPT achieves a high precision and recall of 76.0\% for contradiction. DeepSeek shows improved performance on the entailment class with a precision of 60.6\% and a recall of 80.0\%. However, it underperforms for neutral relations with a low recall of 40.0\%. DeepSeek excels in contradiction prediction, achieving an 81.5\% precision and an 88.0\% recall. ChatGPT demonstrates a higher F1-Score (54.2\%) than DeepSeek (50.0\%) for the neutral class. In contrast, DeepSeek exhibits a better performance in terms of F1-Score than ChatGPT in terms of entailments and instances of contradiction. 

Textual entailment tasks assess the ability of ChatGPT and DeepSeek to determine whether a hypothesis logically follows from a given premise. Evaluations using the SciTail (Table \ref{TE-D17}) and FraCaS (Table \ref{TE-D18}) datasets reveal that DeepSeek outperforms ChatGPT in overall accuracy (64.18\% vs. 62.00\%). On SciTail, a binary entailment classification task, DeepSeek achieves an accuracy of 64.0\% compared to ChatGPT's 60.0\%, showing stronger performance in correctly identifying entailment relationships. On FraCaS, which introduces an additional "contradiction" class, DeepSeek demonstrates better classification balance, particularly excelling in detecting contradictions (81.5\% precision vs. 76.0\% for ChatGPT). Hence, DeepSeek is superior in distinguishing contradictions due to better context comprehension. Both models exhibit weaknesses in handling highly nuanced entailment cases, indicating potential for improvement in logical reasoning tasks. Overall, DeepSeek shows better performance, with an accuracy improvement of 7.50\%.

\subsection{Discussion and Comparative Insights}

This section provides a comprehensive discussion of the performance of ChatGPT and DeepSeek across the five evaluated tasks: sentiment analysis, topic classification, text summarization, machine translation, and textual entailment. The analysis highlights key trends, strengths, and challenges each LLM faces, offering comparative insights that reveal their respective capabilities. Table \ref{sumClas} and Table \ref{sumGen} summarize the average performance per task and the percentage improvement achieved by DeepSeek. The mean results show that ChatGPT outperforms DeepSeek in three out of five NLP tasks.

\begin{table}[h!]
\caption{Summary of ChatGPT and DeepSeek Results on Text Classification Tasks}
\label{sumClas}
\centering
\begin{tabular}{|cc|c|c|c|c|}
\hline
\multicolumn{1}{|c|}{Task}                                                                             & LLM      & Precision & Recall  & F1-Score & Accuracy \\ \hline
\multicolumn{1}{|c|}{\multirow{2}{*}{\begin{tabular}[c]{@{}c@{}}Sentiment \\ Analysis\end{tabular}}}   & ChatGPT  & 75.28\%   & 74.00\% & 74.00\%  & 76.30\%  \\ \cline{2-6} 
\multicolumn{1}{|c|}{}                                                                                 & DeepSeek & 85.54\%   & 85.20\% & 85.22\%  & 87.50\%  \\ \hline
\multicolumn{2}{|c|}{Improvements}                                                                                & 13.63\%   & 15.14\% & 15.17\%  & 14.68\%  \\ \hline
\multicolumn{1}{|c|}{\multirow{2}{*}{\begin{tabular}[c]{@{}c@{}}Topic \\ Classification\end{tabular}}} & ChatGPT  & 60.06\%   & 58.21\% & 57.27\%  & 68.00\%  \\ \cline{2-6} 
\multicolumn{1}{|c|}{}                                                                                 & DeepSeek & 53.41\%   & 54.18\% & 53.14\%  & 66.65\%  \\ \hline
\multicolumn{2}{|c|}{Improvements}                                                                                & -11.08\%  & -6.92\% & -7.22\%  & -1.99\%  \\ \hline
\multicolumn{1}{|c|}{\multirow{2}{*}{\begin{tabular}[c]{@{}c@{}}Textual \\ Entailment\end{tabular}}}   & ChatGPT  & 69.48\%   & 62.40\% & 59.28\%  & 62.00\%  \\ \cline{2-6} 
\multicolumn{1}{|c|}{}                                                                                 & DeepSeek & 73.38\%   & 67.2\% & 64.18\%  & 66.65\%  \\ \hline
\multicolumn{2}{|c|}{Improvements}                                                                                & 5.62\%    & 7.7\%  & 8.27\%   & 7.5\%   \\ \hline
\end{tabular}
\end{table}

\begin{table}[h!]
\caption{Summary of ChatGPT and DeepSeek Results on Text Generation Tasks}
\label{sumGen}
\centering
\begin{tabular}{|cc|ccc|}
\hline
\multicolumn{1}{|c|}{\multirow{2}{*}{Task}}                                                           & \multirow{2}{*}{LLM} & \multicolumn{3}{c|}{BERT Score}                                           \\ \cline{3-5} 
\multicolumn{1}{|c|}{}                                                                                &                      & \multicolumn{1}{c|}{Precision} & \multicolumn{1}{c|}{Recall}   & F1-Score \\ \hline
\multicolumn{1}{|c|}{\multirow{2}{*}{\begin{tabular}[c]{@{}c@{}}Text \\ Summarization\end{tabular}}}  & ChatGPT              & \multicolumn{1}{c|}{69.90\%}   & \multicolumn{1}{c|}{72.94\%}  & 71.34\%  \\ \cline{2-5} 
\multicolumn{1}{|c|}{}                                                                                & DeepSeek             & \multicolumn{1}{c|}{69.53\%}   & \multicolumn{1}{c|}{72.89\%}  & 71.12\%  \\ \hline
\multicolumn{2}{|c|}{Improvements}                                                                                           & \multicolumn{1}{c|}{-0.53\%}   & \multicolumn{1}{c|}{-0.07\%}  & -0.31\%  \\ \hline
\multicolumn{1}{|c|}{\multirow{2}{*}{\begin{tabular}[c]{@{}c@{}}Machine \\ Translation\end{tabular}}} & ChatGPT              & \multicolumn{1}{c|}{78.29\%}   & \multicolumn{1}{c|}{78.49\%}  & 78.36\%  \\ \cline{2-5} 
\multicolumn{1}{|c|}{}                                                                                & DeepSeek             & \multicolumn{1}{c|}{78.30\%}   & \multicolumn{1}{c|}{77.96\%}  & 78.10\%  \\ \hline
\multicolumn{2}{|c|}{Improvements}                                                                                           & \multicolumn{1}{c|}{0.0128\%}   & \multicolumn{1}{c|}{-0.68\%} & -0.33\%  \\ \hline
\end{tabular}
\end{table}

ChatGPT and DeepSeek exhibit distinct performance patterns when dealing with sentiment classification, particularly in a neutral sentiment class. DeepSeek outperforms ChatGPT in overall accuracy, achieving 76.0\% compared to ChatGPT’s 64.7\% (Table \ref{SA-D2}). The confusion matrices reveal that ChatGPT struggles with neutral sentiment, frequently misclassifying neutral samples as positive or negative. In contrast, DeepSeek maintains a more stable classification, demonstrating fewer errors in identifying neutral sentiment, contributing to its superior performance. DeepSeek is generally more balanced across all sentiment classes, leading to fewer extreme misclassifications. Overall, DeepSeek outperforms ChatGPT significantly in all metrics, with a 14.68\% improvement in accuracy.

Topic classification remains challenging for both LLMs, particularly in distinguishing closely related categories. Both models exhibited difficulties with overlapping topics, such as Medical and Psychology, especially in the Web of Science dataset (Table \ref{TC-D20}). Similarly, in the AG News dataset (Table \ref{TC-D19}), ChatGPT faces difficulties differentiating between Business and World News due to semantic similarities. Hence, these findings suggest a need for fine-tuning or incorporating additional context-aware mechanisms. Overall, ChatGPT demonstrates superior precision and accuracy across most categories, making it more suitable for technical fields like CS and Biochemistry. However, DeepSeek performs better for domains requiring high recall, such as Medical and Civil Engineering. Both models require further improvement in niche areas like MAE and Psychology. Additionally, they struggle with ambiguous topics, leading to higher misclassification rates. In this task, ChatGPT outperforms DeepSeek, achieving an accuracy of 68.00\% compared to 66.65\%, reflecting a 1.99\% improvement.

The text summarization task assesses how well each model generates concise and informative summaries. Performance assessments on the Gigaword (Table \ref{TS-D14}) and CNN/Daily Mail (Table \ref{TS-D13}) datasets reveal a mixed trend. On the Gigaword dataset, ChatGPT marginally outperforms DeepSeek across all BERT Score metrics, achieving a precision of 71.00\% and an F1-Score of 71.59\%, compared to 70.62\% precision and 71.11\% F1-Score for DeepSeek. Conversely, on CNN/Daily Mail dataset, DeepSeek slightly outperforms ChatGPT in F1-Score (71.13\% vs. 71.09\%), but ChatGPT achieves slightly higher precision. The results indicate that ChatGPT generates more concise summaries, contributing to higher precision. DeepSeek demonstrates superior recall, suggesting a broader content capture. However, the differences between the LLMs are minor, indicating both are effective summarization tools, with trade-offs depending on whether conciseness (ChatGPT) or content coverage (DeepSeek) is prioritized. Overall, ChatGPT outperforms DeepSeek by 0.31\% in this task.

The machine translation task evaluates how well ChatGPT and DeepSeek handle English-to-Arabic translations using the ArzEn-MultiGenre (Table \ref{MT-D7}) and AraBench (Table \ref{MT-D6}) datasets. The results reveal that performance varies based on the dialect of Arabic used. On ArzEn-MultiGenre, which includes Egyptian Arabic song lyrics, novels, and subtitles, ChatGPT outperforms DeepSeek across all BERT Score metrics, achieving an F1-Score of 78.39\% vs. 77.53\% for DeepSeek. On AraBench, which includes Jordanian and Qatari dialects, DeepSeek slightly outperforms ChatGPT, achieving an F1-Score of 78.67\% vs. 78.33\% for ChatGPT. Despite these minor variations, both models show similar overall performance, with slight differences based on the dataset and the Arabic dialect. In general, ChatGPT demonstrates stronger performance in machine translation tasks compared to DeepSeek, with an 0.3318\% improvement.

To recap, DeepSeek demonstrates a stronger performance in structured tasks, such as sentiment analysis and textual entailment, where its classification stability is likely a contributing factor. The model consistently exhibits reliable results in these tasks, indicating its robustness in handling structured data with clear categories. On the other hand, ChatGPT excels in more subjective and nuanced tasks, such as topic classification, summarization, and certain translation cases. These findings indicate that ChatGPT may possess a more refined understanding of context, enabling it to perform well in tasks that require a deeper interpretation of meaning and subtleties.

Both models perform similarly regarding text summarization and translation, with the trade-off between precision and recall being a key differentiating factor. ChatGPT tends to prioritize conciseness and precision, while DeepSeek may capture more comprehensive content, showing better recall. These findings indicate that the choice between the models may depend on the specific requirements of the task, such as whether precision or recall is more critical.

Despite these individual strengths, neither model consistently outperforms the other across all tasks. Their applicability depends on the nature of the task, suggesting that one model may be more suitable than the other depending on specific requirements. Moreover, improvements are needed in handling tasks that involve neutrality and contradictions. Specifically, ChatGPT shows room for improvement in sentiment analysis and textual entailment tasks, where it struggles with neutral classifications and contradictions.

\section{Conclusion}
\label{con}

This study evaluated the performance of ChatGPT and DeepSeek across multiple NLP tasks, including sentiment analysis, topic classification, text summarization, machine translation, and textual entailment. The methodology employed in this study ensured a systematic and unbiased evaluation of both LLMs. A well-defined experimental protocol governed the querying process, response collection, and evaluation framework, minimizing variability sources and ensuring comparison fairness. Identical prompts are used for both LLMs and are designed to be neutral, clear, and representative of real-world use cases. For tasks like textual entailment and machine translation, few-shot examples are incorporated to evaluate the models' contextual responses. Two benchmark datasets are selected for each NLP task, ensuring a comprehensive evaluation across diverse domains, including news, reviews, and formal/informal texts. 

The results indicate that both LLMs exhibit strengths and weaknesses depending on the nature of the task. DeepSeek generally excels in structured tasks, such as sentiment analysis and textual entailment, where its classification stability and logical reasoning abilities are particularly evident. In contrast, ChatGPT demonstrates superior performance in more subjective and nuanced tasks, including topic classification, summarization, and translation, where its refined contextual understanding is particularly valuable. Both models show similar performance in summarization and translation, with trade-offs between precision and recall, making their suitability task-dependent. DeepSeek favors recall, ensuring comprehensive content capture, whereas ChatGPT prioritizes precision, leading to more concise outputs. However, both LLMs have areas for improvement. Specifically, handling contradictions, neutrality, and domain-specific classifications remains challenging. These weaknesses point to the need for further advancements in logical reasoning and nuanced content understanding for both models.

\bibliography{references}

\begin{thebibliography}{45}
\expandafter\ifx\csname natexlab\endcsname\relax\def\natexlab#1{#1}\fi
\providecommand{\url}[1]{\texttt{#1}}
\providecommand{\href}[2]{#2}
\providecommand{\path}[1]{#1}
\providecommand{\DOIprefix}{doi:}
\providecommand{\ArXivprefix}{arXiv:}
\providecommand{\URLprefix}{URL: }
\providecommand{\Pubmedprefix}{pmid:}
\providecommand{\doi}[1]{\href{http://dx.doi.org/#1}{\path{#1}}}
\providecommand{\Pubmed}[1]{\href{pmid:#1}{\path{#1}}}
\providecommand{\bibinfo}[2]{#2}
\ifx\xfnm\relax \def\xfnm[#1]{\unskip,\space#1}\fi
\bibitem[{Achiam et~al.(2023)Achiam, Adler, Agarwal, Ahmad, Akkaya, Aleman, Almeida, Altenschmidt, Altman, Anadkat et~al.}]{achiam2023gpt}
\bibinfo{author}{Achiam, J.}, \bibinfo{author}{Adler, S.}, \bibinfo{author}{Agarwal, S.}, \bibinfo{author}{Ahmad, L.}, \bibinfo{author}{Akkaya, I.}, \bibinfo{author}{Aleman, F.~L.}, \bibinfo{author}{Almeida, D.}, \bibinfo{author}{Altenschmidt, J.}, \bibinfo{author}{Altman, S.}, \bibinfo{author}{Anadkat, S.} et~al. (\bibinfo{year}{2023}).
\newblock \bibinfo{title}{Gpt-4 technical report}.
\newblock {\it \bibinfo{journal}{arXiv preprint arXiv:2303.08774}\/}, .
\bibitem[{Al-Sabbagh(2024)}]{AlSabbagh2024}
\bibinfo{author}{Al-Sabbagh, R.} (\bibinfo{year}{2024}).
\newblock \bibinfo{title}{Arzen-multigenre: An aligned parallel dataset of egyptian arabic song lyrics, novels, and subtitles, with english translations}.
\newblock {\it \bibinfo{journal}{Data in Brief}\/},  {\it \bibinfo{volume}{54}\/}, \bibinfo{pages}{110271}. \URLprefix \url{http://dx.doi.org/10.1016/j.dib.2024.110271}. \DOIprefix\doi{10.1016/j.dib.2024.110271}.
\bibitem[{Antaki et~al.(2023)Antaki, Touma, Milad, El-Khoury \& Duval}]{antaki2023evaluating}
\bibinfo{author}{Antaki, F.}, \bibinfo{author}{Touma, S.}, \bibinfo{author}{Milad, D.}, \bibinfo{author}{El-Khoury, J.}, \& \bibinfo{author}{Duval, R.} (\bibinfo{year}{2023}).
\newblock \bibinfo{title}{Evaluating the performance of chatgpt in ophthalmology: an analysis of its successes and shortcomings}.
\newblock {\it \bibinfo{journal}{Ophthalmology science}\/},  {\it \bibinfo{volume}{3}\/}, \bibinfo{pages}{100324}.
\bibitem[{Augenstein et~al.(2017)Augenstein, Das, Riedel, Vikraman \& McCallum}]{augenstein-etal-2017-semeval}
\bibinfo{author}{Augenstein, I.}, \bibinfo{author}{Das, M.}, \bibinfo{author}{Riedel, S.}, \bibinfo{author}{Vikraman, L.}, \& \bibinfo{author}{McCallum, A.} (\bibinfo{year}{2017}).
\newblock \bibinfo{title}{{S}em{E}val 2017 task 10: {S}cience{IE} - extracting keyphrases and relations from scientific publications}.
\newblock In \bibinfo{editor}{S.~Bethard}, \bibinfo{editor}{M.~Carpuat}, \bibinfo{editor}{M.~Apidianaki}, \bibinfo{editor}{S.~M. Mohammad}, \bibinfo{editor}{D.~Cer}, \& \bibinfo{editor}{D.~Jurgens} (Eds.), {\it \bibinfo{booktitle}{Proceedings of the 11th International Workshop on Semantic Evaluation ({S}em{E}val-2017)}\/} (pp. \bibinfo{pages}{546--555}).
\newblock \bibinfo{address}{Vancouver, Canada}: \bibinfo{publisher}{Association for Computational Linguistics}.
\newblock \URLprefix \url{https://aclanthology.org/S17-2091/}. \DOIprefix\doi{10.18653/v1/S17-2091}.
\bibitem[{Bahrini et~al.(2023)Bahrini, Khamoshifar, Abbasimehr, Riggs, Esmaeili, Majdabadkohne \& Pasehvar}]{bahrini2023chatgpt}
\bibinfo{author}{Bahrini, A.}, \bibinfo{author}{Khamoshifar, M.}, \bibinfo{author}{Abbasimehr, H.}, \bibinfo{author}{Riggs, R.~J.}, \bibinfo{author}{Esmaeili, M.}, \bibinfo{author}{Majdabadkohne, R.~M.}, \& \bibinfo{author}{Pasehvar, M.} (\bibinfo{year}{2023}).
\newblock \bibinfo{title}{Chatgpt: Applications, opportunities, and threats}.
\newblock In {\it \bibinfo{booktitle}{2023 Systems and Information Engineering Design Symposium (SIEDS)}\/} (pp. \bibinfo{pages}{274--279}).
\newblock \bibinfo{organization}{IEEE}.
\bibitem[{Biltawi et~al.(2016)Biltawi, Etaiwi, Tedmori, Hudaib \& Awajan}]{Biltawi7476075}
\bibinfo{author}{Biltawi, M.}, \bibinfo{author}{Etaiwi, W.}, \bibinfo{author}{Tedmori, S.}, \bibinfo{author}{Hudaib, A.}, \& \bibinfo{author}{Awajan, A.} (\bibinfo{year}{2016}).
\newblock \bibinfo{title}{Sentiment classification techniques for arabic language: A survey}.
\newblock In {\it \bibinfo{booktitle}{2016 7th International Conference on Information and Communication Systems (ICICS)}\/} (pp. \bibinfo{pages}{339--346}).
\newblock \DOIprefix\doi{10.1109/IACS.2016.7476075}.
\bibitem[{Caramancion(2023)}]{caramancion2023news}
\bibinfo{author}{Caramancion, K.~M.} (\bibinfo{year}{2023}).
\newblock \bibinfo{title}{News verifiers showdown: a comparative performance evaluation of chatgpt 3.5, chatgpt 4.0, bing ai, and bard in news fact-checking}.
\newblock In {\it \bibinfo{booktitle}{2023 IEEE Future Networks World Forum (FNWF)}\/} (pp. \bibinfo{pages}{1--6}).
\newblock \bibinfo{organization}{IEEE}.
\bibitem[{Chen et~al.(2024)Chen, Wang, Yu \& Zhou}]{chen2024artificial}
\bibinfo{author}{Chen, Y.}, \bibinfo{author}{Wang, H.}, \bibinfo{author}{Yu, K.}, \& \bibinfo{author}{Zhou, R.} (\bibinfo{year}{2024}).
\newblock \bibinfo{title}{Artificial intelligence methods in natural language processing: A comprehensive review}.
\newblock {\it \bibinfo{journal}{Highlights in Science Engineering and Technology}\/},  {\it \bibinfo{volume}{85}\/}, \bibinfo{pages}{545--550}.
\bibitem[{Coello et~al.(2024)Coello, Alimam \& Kouatly}]{Coello2024}
\bibinfo{author}{Coello, C. E.~A.}, \bibinfo{author}{Alimam, M.~N.}, \& \bibinfo{author}{Kouatly, R.} (\bibinfo{year}{2024}).
\newblock \bibinfo{title}{Effectiveness of chatgpt in coding: A comparative analysis of popular large language models}.
\newblock {\it \bibinfo{journal}{Digital}\/},  {\it \bibinfo{volume}{4}\/}, \bibinfo{pages}{114–125}. \URLprefix \url{http://dx.doi.org/10.3390/digital4010005}. \DOIprefix\doi{10.3390/digital4010005}.
\bibitem[{Cooper et~al.(1996)Cooper, Crouch, Van~Eijck, Fox, Van~Genabith, Jaspars, Kamp, Milward, Pinkal, Poesio et~al.}]{cooper1996using}
\bibinfo{author}{Cooper, R.}, \bibinfo{author}{Crouch, D.}, \bibinfo{author}{Van~Eijck, J.}, \bibinfo{author}{Fox, C.}, \bibinfo{author}{Van~Genabith, J.}, \bibinfo{author}{Jaspars, J.}, \bibinfo{author}{Kamp, H.}, \bibinfo{author}{Milward, D.}, \bibinfo{author}{Pinkal, M.}, \bibinfo{author}{Poesio, M.} et~al. (\bibinfo{year}{1996}).
\newblock {\it \bibinfo{title}{Using the framework}\/}.
\newblock \bibinfo{type}{Technical Report} Technical Report LRE 62-051 D-16, The FraCaS Consortium.
\bibitem[{Deandres-Tame et~al.(2024)Deandres-Tame, Tolosana, Vera-Rodriguez, Morales, Fierrez \& Ortega-Garcia}]{deandres2024good}
\bibinfo{author}{Deandres-Tame, I.}, \bibinfo{author}{Tolosana, R.}, \bibinfo{author}{Vera-Rodriguez, R.}, \bibinfo{author}{Morales, A.}, \bibinfo{author}{Fierrez, J.}, \& \bibinfo{author}{Ortega-Garcia, J.} (\bibinfo{year}{2024}).
\newblock \bibinfo{title}{How good is chatgpt at face biometrics? a first look into recognition, soft biometrics, and explainability}.
\newblock {\it \bibinfo{journal}{IEEE Access}\/}, .
\bibitem[{Del~Corso et~al.(2005)Del~Corso, Gull\'{\i} \& Romani}]{Corso2005}
\bibinfo{author}{Del~Corso, G.~M.}, \bibinfo{author}{Gull\'{\i}, A.}, \& \bibinfo{author}{Romani, F.} (\bibinfo{year}{2005}).
\newblock \bibinfo{title}{Ranking a stream of news}.
\newblock In {\it \bibinfo{booktitle}{Proceedings of the 14th International Conference on World Wide Web}\/} WWW '05 (p. \bibinfo{pages}{97–106}).
\newblock \bibinfo{address}{New York, NY, USA}: \bibinfo{publisher}{Association for Computing Machinery}.
\newblock \URLprefix \url{https://doi.org/10.1145/1060745.1060764}. \DOIprefix\doi{10.1145/1060745.1060764}.
\bibitem[{Dunder et~al.(2024)Dunder, Lundborg, Wong \& Viberg}]{dunder2024kattis}
\bibinfo{author}{Dunder, N.}, \bibinfo{author}{Lundborg, S.}, \bibinfo{author}{Wong, J.}, \& \bibinfo{author}{Viberg, O.} (\bibinfo{year}{2024}).
\newblock \bibinfo{title}{Kattis vs chatgpt: Assessment and evaluation of programming tasks in the age of artificial intelligence}.
\newblock In {\it \bibinfo{booktitle}{Proceedings of the 14th Learning Analytics and Knowledge Conference}\/} (pp. \bibinfo{pages}{821--827}).
\bibitem[{Elyoseph et~al.(2023)Elyoseph, Hadar-Shoval, Asraf \& Lvovsky}]{elyoseph2023chatgpt}
\bibinfo{author}{Elyoseph, Z.}, \bibinfo{author}{Hadar-Shoval, D.}, \bibinfo{author}{Asraf, K.}, \& \bibinfo{author}{Lvovsky, M.} (\bibinfo{year}{2023}).
\newblock \bibinfo{title}{Chatgpt outperforms humans in emotional awareness evaluations}.
\newblock {\it \bibinfo{journal}{Frontiers in psychology}\/},  {\it \bibinfo{volume}{14}\/}, \bibinfo{pages}{1199058}.
\bibitem[{Feuerriegel et~al.(2024)Feuerriegel, Hartmann, Janiesch \& Zschech}]{feuerriegel2024generative}
\bibinfo{author}{Feuerriegel, S.}, \bibinfo{author}{Hartmann, J.}, \bibinfo{author}{Janiesch, C.}, \& \bibinfo{author}{Zschech, P.} (\bibinfo{year}{2024}).
\newblock \bibinfo{title}{Generative ai}.
\newblock {\it \bibinfo{journal}{Business \& Information Systems Engineering}\/},  {\it \bibinfo{volume}{66}\/}, \bibinfo{pages}{111--126}.
\bibitem[{Fidelangeli et~al.(2025)Fidelangeli, Galli, Loreggia, Pisano, Rovatti, Santin \& Sartor}]{Fidelangeli2025}
\bibinfo{author}{Fidelangeli, A.}, \bibinfo{author}{Galli, F.}, \bibinfo{author}{Loreggia, A.}, \bibinfo{author}{Pisano, G.}, \bibinfo{author}{Rovatti, R.}, \bibinfo{author}{Santin, P.}, \& \bibinfo{author}{Sartor, G.} (\bibinfo{year}{2025}).
\newblock \bibinfo{title}{The summarization of italian tax-law decisions: The case of the prodigit project}.
\newblock {\it \bibinfo{journal}{IEEE Access}\/},  {\it \bibinfo{volume}{13}\/}, \bibinfo{pages}{38833–38855}. \URLprefix \url{http://dx.doi.org/10.1109/ACCESS.2025.3545419}. \DOIprefix\doi{10.1109/access.2025.3545419}.
\bibitem[{Fu et~al.(2024)Fu, Wang \& Li}]{fu2024can}
\bibinfo{author}{Fu, X.}, \bibinfo{author}{Wang, R.}, \& \bibinfo{author}{Li, C.} (\bibinfo{year}{2024}).
\newblock \bibinfo{title}{Can chatgpt evaluate plans?}
\newblock {\it \bibinfo{journal}{Journal of the American Planning Association}\/},  {\it \bibinfo{volume}{90}\/}, \bibinfo{pages}{525--536}.
\bibitem[{González et~al.(2024)González, Poenaru, Woo, Trappey, Carter, Darcy, Encisco, Gulack, Miniati, Tombash \& Huang}]{Gonzlez2024}
\bibinfo{author}{González, R.}, \bibinfo{author}{Poenaru, D.}, \bibinfo{author}{Woo, R.}, \bibinfo{author}{Trappey, A.~F.}, \bibinfo{author}{Carter, S.}, \bibinfo{author}{Darcy, D.}, \bibinfo{author}{Encisco, E.}, \bibinfo{author}{Gulack, B.}, \bibinfo{author}{Miniati, D.}, \bibinfo{author}{Tombash, E.}, \& \bibinfo{author}{Huang, E.~Y.} (\bibinfo{year}{2024}).
\newblock \bibinfo{title}{Chatgpt: What every pediatric surgeon should know about its potential uses and pitfalls}.
\newblock {\it \bibinfo{journal}{Journal of Pediatric Surgery}\/},  {\it \bibinfo{volume}{59}\/}, \bibinfo{pages}{941–947}. \URLprefix \url{http://dx.doi.org/10.1016/j.jpedsurg.2024.01.007}. \DOIprefix\doi{10.1016/j.jpedsurg.2024.01.007}.
\bibitem[{Islam et~al.(2023)Islam, Elmekki, Elsebai, Bentahar, Drawel, Rjoub \& Pedrycz}]{islam2023comprehensive}
\bibinfo{author}{Islam, S.}, \bibinfo{author}{Elmekki, H.}, \bibinfo{author}{Elsebai, A.}, \bibinfo{author}{Bentahar, J.}, \bibinfo{author}{Drawel, N.}, \bibinfo{author}{Rjoub, G.}, \& \bibinfo{author}{Pedrycz, W.} (\bibinfo{year}{2023}).
\newblock \bibinfo{title}{A comprehensive survey on applications of transformers for deep learning tasks}.
\newblock {\it \bibinfo{journal}{Expert Systems with Applications}\/},  (p. \bibinfo{pages}{122666}).
\bibitem[{Jiang et~al.(2025)Jiang, Gao \& Karniadakis}]{Jiang2025}
\bibinfo{author}{Jiang, Q.}, \bibinfo{author}{Gao, Z.}, \& \bibinfo{author}{Karniadakis, G.~E.} (\bibinfo{year}{2025}).
\newblock \bibinfo{title}{Deepseek vs. chatgpt vs. claude: A comparative study for scientific computing and scientific machine learning tasks}.
\newblock {\it \bibinfo{journal}{Theoretical and Applied Mechanics Letters}\/},  (p. \bibinfo{pages}{100583}). \URLprefix \url{http://dx.doi.org/10.1016/j.taml.2025.100583}. \DOIprefix\doi{10.1016/j.taml.2025.100583}.
\bibitem[{Johnson et~al.(2023)Johnson, Goodman, Patrinely, Stone, Zimmerman, Donald, Chang, Berkowitz, Finn, Jahangir et~al.}]{johnson2023assessing}
\bibinfo{author}{Johnson, D.}, \bibinfo{author}{Goodman, R.}, \bibinfo{author}{Patrinely, J.}, \bibinfo{author}{Stone, C.}, \bibinfo{author}{Zimmerman, E.}, \bibinfo{author}{Donald, R.}, \bibinfo{author}{Chang, S.}, \bibinfo{author}{Berkowitz, S.}, \bibinfo{author}{Finn, A.}, \bibinfo{author}{Jahangir, E.} et~al. (\bibinfo{year}{2023}).
\newblock \bibinfo{title}{Assessing the accuracy and reliability of ai-generated medical responses: an evaluation of the chat-gpt model}.
\newblock {\it \bibinfo{journal}{Research square}\/},  (pp. \bibinfo{pages}{rs--3}).
\bibitem[{Khlaif et~al.(2023)Khlaif, Mousa, Hattab, Itmazi, Hassan, Sanmugam \& Ayyoub}]{khlaif2023potential}
\bibinfo{author}{Khlaif, Z.~N.}, \bibinfo{author}{Mousa, A.}, \bibinfo{author}{Hattab, M.~K.}, \bibinfo{author}{Itmazi, J.}, \bibinfo{author}{Hassan, A.~A.}, \bibinfo{author}{Sanmugam, M.}, \& \bibinfo{author}{Ayyoub, A.} (\bibinfo{year}{2023}).
\newblock \bibinfo{title}{The potential and concerns of using ai in scientific research: Chatgpt performance evaluation}.
\newblock {\it \bibinfo{journal}{JMIR Medical Education}\/},  {\it \bibinfo{volume}{9}\/}, \bibinfo{pages}{e47049}.
\bibitem[{Khot et~al.(2018)Khot, Sabharwal \& Clark}]{Khot2018}
\bibinfo{author}{Khot, T.}, \bibinfo{author}{Sabharwal, A.}, \& \bibinfo{author}{Clark, P.} (\bibinfo{year}{2018}).
\newblock \bibinfo{title}{Scitail: A textual entailment dataset from science question answering}.
\newblock {\it \bibinfo{journal}{Proceedings of the AAAI Conference on Artificial Intelligence}\/},  {\it \bibinfo{volume}{32}\/}. \URLprefix \url{http://dx.doi.org/10.1609/aaai.v32i1.12022}. \DOIprefix\doi{10.1609/aaai.v32i1.12022}.
\bibitem[{Kowsari(2018)}]{hKowsari2018}
\bibinfo{author}{Kowsari, K.} (\bibinfo{year}{2018}).
\newblock \bibinfo{title}{Web of science dataset}.
\newblock \URLprefix \url{https://data.mendeley.com/datasets/9rw3vkcfy4/6}. \DOIprefix\doi{10.17632/9RW3VKCFY4.6}.
\bibitem[{Kowsari et~al.(2017)Kowsari, Brown, Heidarysafa, Jafari~Meimandi, , Gerber \& Barnes}]{kowsari2017HDLTex}
\bibinfo{author}{Kowsari, K.}, \bibinfo{author}{Brown, D.~E.}, \bibinfo{author}{Heidarysafa, M.}, \bibinfo{author}{Jafari~Meimandi, K.}, , \bibinfo{author}{Gerber, M.~S.}, \& \bibinfo{author}{Barnes, L.~E.} (\bibinfo{year}{2017}).
\newblock \bibinfo{title}{Hdltex: Hierarchical deep learning for text classification}.
\newblock In {\it \bibinfo{booktitle}{Machine Learning and Applications (ICMLA), 2017 16th IEEE International Conference on}\/}.
\newblock \bibinfo{organization}{IEEE}.
\bibitem[{Kumar(2024)}]{kumar2024large}
\bibinfo{author}{Kumar, P.} (\bibinfo{year}{2024}).
\newblock \bibinfo{title}{Large language models (llms): survey, technical frameworks, and future challenges}.
\newblock {\it \bibinfo{journal}{Artificial Intelligence Review}\/},  {\it \bibinfo{volume}{57}\/}, \bibinfo{pages}{260}.
\bibitem[{Li et~al.(2024)Li, Fan, Gu, Li, Duan, Dong, Liu \& Wang}]{li2024flexkbqa}
\bibinfo{author}{Li, Z.}, \bibinfo{author}{Fan, S.}, \bibinfo{author}{Gu, Y.}, \bibinfo{author}{Li, X.}, \bibinfo{author}{Duan, Z.}, \bibinfo{author}{Dong, B.}, \bibinfo{author}{Liu, N.}, \& \bibinfo{author}{Wang, J.} (\bibinfo{year}{2024}).
\newblock \bibinfo{title}{Flexkbqa: A flexible llm-powered framework for few-shot knowledge base question answering}.
\newblock In {\it \bibinfo{booktitle}{Proceedings of the AAAI conference on artificial intelligence}\/} (pp. \bibinfo{pages}{18608--18616}).
\newblock volume~\bibinfo{volume}{38}.
\bibitem[{Liao(2025)}]{liao2025deepseek}
\bibinfo{author}{Liao, H.} (\bibinfo{year}{2025}).
\newblock \bibinfo{title}{Deepseek large-scale model: technical analysis and development prospect}.
\newblock {\it \bibinfo{journal}{Journal of Computer Science and Electrical Engineering}\/},  {\it \bibinfo{volume}{7}\/}, \bibinfo{pages}{33--37}.
\bibitem[{Liu et~al.(2024{\natexlab{a}})Liu, Feng, Xue, Wang, Wu, Lu, Zhao, Deng, Zhang, Ruan et~al.}]{liu2024deepseek}
\bibinfo{author}{Liu, A.}, \bibinfo{author}{Feng, B.}, \bibinfo{author}{Xue, B.}, \bibinfo{author}{Wang, B.}, \bibinfo{author}{Wu, B.}, \bibinfo{author}{Lu, C.}, \bibinfo{author}{Zhao, C.}, \bibinfo{author}{Deng, C.}, \bibinfo{author}{Zhang, C.}, \bibinfo{author}{Ruan, C.} et~al. (\bibinfo{year}{2024}{\natexlab{a}}).
\newblock \bibinfo{title}{Deepseek-v3 technical report}.
\newblock {\it \bibinfo{journal}{arXiv preprint arXiv:2412.19437}\/}, .
\bibitem[{Liu et~al.(2024{\natexlab{b}})Liu, Xie, Zhao, Zhou, Xu, Li \& Chen}]{liu2024speak}
\bibinfo{author}{Liu, C.}, \bibinfo{author}{Xie, Z.}, \bibinfo{author}{Zhao, S.}, \bibinfo{author}{Zhou, J.}, \bibinfo{author}{Xu, T.}, \bibinfo{author}{Li, M.}, \& \bibinfo{author}{Chen, E.} (\bibinfo{year}{2024}{\natexlab{b}}).
\newblock \bibinfo{title}{Speak from heart: an emotion-guided llm-based multimodal method for emotional dialogue generation}.
\newblock In {\it \bibinfo{booktitle}{Proceedings of the 2024 International Conference on Multimedia Retrieval}\/} (pp. \bibinfo{pages}{533--542}).
\bibitem[{Liu et~al.(2024{\natexlab{c}})Liu, Zhu, Pang, Xue, Zhang \& Fan}]{Liu2024}
\bibinfo{author}{Liu, X.}, \bibinfo{author}{Zhu, Y.}, \bibinfo{author}{Pang, T.}, \bibinfo{author}{Xue, K.}, \bibinfo{author}{Zhang, X.}, \& \bibinfo{author}{Fan, C.} (\bibinfo{year}{2024}{\natexlab{c}}).
\newblock \bibinfo{title}{Medical document embedding enhancement with heterogeneous mixture-of-experts}.
\newblock In {\it \bibinfo{booktitle}{2024 IEEE International Conference on Bioinformatics and Biomedicine (BIBM)}\/} (pp. \bibinfo{pages}{2238--2243}).
\newblock \DOIprefix\doi{10.1109/BIBM62325.2024.10822374}.
\bibitem[{Maas et~al.(2011)Maas, Daly, Pham, Huang, Ng \& Potts}]{maasIMDB2011}
\bibinfo{author}{Maas, A.~L.}, \bibinfo{author}{Daly, R.~E.}, \bibinfo{author}{Pham, P.~T.}, \bibinfo{author}{Huang, D.}, \bibinfo{author}{Ng, A.~Y.}, \& \bibinfo{author}{Potts, C.} (\bibinfo{year}{2011}).
\newblock \bibinfo{title}{Learning word vectors for sentiment analysis}.
\newblock In {\it \bibinfo{booktitle}{Proceedings of the 49th Annual Meeting of the Association for Computational Linguistics: Human Language Technologies}\/} (pp. \bibinfo{pages}{142--150}).
\newblock \bibinfo{address}{Portland, Oregon, USA}: \bibinfo{publisher}{Association for Computational Linguistics}.
\newblock \URLprefix \url{http://www.aclweb.org/anthology/P11-1015}.
\bibitem[{Mo \& Wu(2025, Access Date: 29/3/2025)}]{Mo2025DeepSeek}
\bibinfo{author}{Mo, L.}, \& \bibinfo{author}{Wu, K.} (\bibinfo{year}{2025, Access Date: 29/3/2025}).
\newblock \bibinfo{title}{Deepseek narrows china-us ai gap to three months, 01.ai founder lee kai-fu says}.
\newblock {\it \bibinfo{journal}{Reuters}\/}, .
\bibitem[{Ouyang et~al.(2022)Ouyang, Wu, Jiang, Almeida, Wainwright, Mishkin, Zhang, Agarwal, Slama, Ray, Schulman, Hilton, Kelton, Miller, Simens, Askell, Welinder, Christiano, Leike \& Lowe}]{Ouyang2022}
\bibinfo{author}{Ouyang, L.}, \bibinfo{author}{Wu, J.}, \bibinfo{author}{Jiang, X.}, \bibinfo{author}{Almeida, D.}, \bibinfo{author}{Wainwright, C.~L.}, \bibinfo{author}{Mishkin, P.}, \bibinfo{author}{Zhang, C.}, \bibinfo{author}{Agarwal, S.}, \bibinfo{author}{Slama, K.}, \bibinfo{author}{Ray, A.}, \bibinfo{author}{Schulman, J.}, \bibinfo{author}{Hilton, J.}, \bibinfo{author}{Kelton, F.}, \bibinfo{author}{Miller, L.}, \bibinfo{author}{Simens, M.}, \bibinfo{author}{Askell, A.}, \bibinfo{author}{Welinder, P.}, \bibinfo{author}{Christiano, P.}, \bibinfo{author}{Leike, J.}, \& \bibinfo{author}{Lowe, R.} (\bibinfo{year}{2022}).
\newblock \bibinfo{title}{Training language models to follow instructions with human feedback}.
\newblock In {\it \bibinfo{booktitle}{Proceedings of the 36th International Conference on Neural Information Processing Systems}\/} NIPS '22.
\newblock \bibinfo{address}{Red Hook, NY, USA}: \bibinfo{publisher}{Curran Associates Inc.}
\bibitem[{Rezaei et~al.(2024)Rezaei, Salehi \& Tabatabaei}]{Rezaei2024}
\bibinfo{author}{Rezaei, M.}, \bibinfo{author}{Salehi, H.}, \& \bibinfo{author}{Tabatabaei, O.} (\bibinfo{year}{2024}).
\newblock \bibinfo{title}{Uses and misuses of chatgpt as an ai-language model in academic writing}.
\newblock In {\it \bibinfo{booktitle}{2024 10th International Conference on Artificial Intelligence and Robotics (QICAR)}\/} (p. \bibinfo{pages}{256–260}).
\newblock \bibinfo{publisher}{IEEE}.
\newblock \URLprefix \url{http://dx.doi.org/10.1109/QICAR61538.2024.10496607}. \DOIprefix\doi{10.1109/qicar61538.2024.10496607}.
\bibitem[{Rush(2024)}]{RushGigaword2024}
\bibinfo{author}{Rush, A.~M.} (\bibinfo{year}{2024}).
\newblock \bibinfo{title}{Gigaword dataset}.
\newblock \URLprefix \url{https://service.tib.eu/ldmservice/dataset/5e1639e1-c107-48c0-80a8-c593c6b8ad5b}. \DOIprefix\doi{10.57702/XAWJFKMB}.
\bibitem[{Sajjad et~al.(2020)Sajjad, Abdelali, Durrani \& Dalvi}]{Sajjad2020}
\bibinfo{author}{Sajjad, H.}, \bibinfo{author}{Abdelali, A.}, \bibinfo{author}{Durrani, N.}, \& \bibinfo{author}{Dalvi, F.} (\bibinfo{year}{2020}).
\newblock \bibinfo{title}{Arabench: Benchmarking dialectal arabic-english machine translation}.
\newblock In {\it \bibinfo{booktitle}{Proceedings of the 28th International Conference on Computational Linguistics}\/} (p. \bibinfo{pages}{5094–5107}).
\newblock \bibinfo{publisher}{International Committee on Computational Linguistics}.
\newblock \URLprefix \url{http://dx.doi.org/10.18653/v1/2020.coling-main.447}. \DOIprefix\doi{10.18653/v1/2020.coling-main.447}.
\bibitem[{van Schaik \& Pugh(2024)}]{van2024field}
\bibinfo{author}{van Schaik, T.~A.}, \& \bibinfo{author}{Pugh, B.} (\bibinfo{year}{2024}).
\newblock \bibinfo{title}{A field guide to automatic evaluation of llm-generated summaries}.
\newblock In {\it \bibinfo{booktitle}{Proceedings of the 47th International ACM SIGIR Conference on Research and Development in Information Retrieval}\/} (pp. \bibinfo{pages}{2832--2836}).
\bibitem[{See et~al.(2017)See, Liu \& Manning}]{see-etal-2017-get}
\bibinfo{author}{See, A.}, \bibinfo{author}{Liu, P.~J.}, \& \bibinfo{author}{Manning, C.~D.} (\bibinfo{year}{2017}).
\newblock \bibinfo{title}{Get to the point: Summarization with pointer-generator networks}.
\newblock In {\it \bibinfo{booktitle}{Proceedings of the 55th Annual Meeting of the Association for Computational Linguistics (Volume 1: Long Papers)}\/} (pp. \bibinfo{pages}{1073--1083}).
\newblock \bibinfo{address}{Vancouver, Canada}: \bibinfo{publisher}{Association for Computational Linguistics}.
\newblock \URLprefix \url{https://www.aclweb.org/anthology/P17-1099}. \DOIprefix\doi{10.18653/v1/P17-1099}.
\bibitem[{Silva et~al.(2015)Silva, Ferreira, Lins, Cabral, Oliveira, Simske \& Riss}]{silva2015automatic}
\bibinfo{author}{Silva, G.}, \bibinfo{author}{Ferreira, R.}, \bibinfo{author}{Lins, R.~D.}, \bibinfo{author}{Cabral, L.}, \bibinfo{author}{Oliveira, H.}, \bibinfo{author}{Simske, S.~J.}, \& \bibinfo{author}{Riss, M.} (\bibinfo{year}{2015}).
\newblock \bibinfo{title}{Automatic text document summarization based on machine learning}.
\newblock In {\it \bibinfo{booktitle}{Proceedings of the 2015 ACM Symposium on Document Engineering}\/} (pp. \bibinfo{pages}{191--194}).
\newblock \bibinfo{organization}{ACM}.
\bibitem[{Sunagar et~al.(2021)Sunagar, Kanavalli, Nayak, Mahan, Prasad \& Prasad}]{Sunagar2021}
\bibinfo{author}{Sunagar, P.}, \bibinfo{author}{Kanavalli, A.}, \bibinfo{author}{Nayak, S.~S.}, \bibinfo{author}{Mahan, S.~R.}, \bibinfo{author}{Prasad, S.}, \& \bibinfo{author}{Prasad, S.} (\bibinfo{year}{2021}).
\newblock \bibinfo{title}{News topic classification using machine learning techniques}.
\newblock In \bibinfo{editor}{V.~Bindhu}, \bibinfo{editor}{J.~M. R.~S. Tavares}, \bibinfo{editor}{A.-A.~A. Boulogeorgos}, \& \bibinfo{editor}{C.~Vuppalapati} (Eds.), {\it \bibinfo{booktitle}{International Conference on Communication, Computing and Electronics Systems}\/} (pp. \bibinfo{pages}{461--474}).
\newblock \bibinfo{address}{Singapore}: \bibinfo{publisher}{Springer Singapore}.
\bibitem[{Thelwall(2024)}]{thelwall2024can}
\bibinfo{author}{Thelwall, M.} (\bibinfo{year}{2024}).
\newblock \bibinfo{title}{Can chatgpt evaluate research quality?}
\newblock {\it \bibinfo{journal}{Journal of Data and Information Science}\/},  {\it \bibinfo{volume}{9}\/}, \bibinfo{pages}{1--21}.
\bibitem[{Wang et~al.(2022)Wang, Wu, He, Huang \& Church}]{Wang2022}
\bibinfo{author}{Wang, H.}, \bibinfo{author}{Wu, H.}, \bibinfo{author}{He, Z.}, \bibinfo{author}{Huang, L.}, \& \bibinfo{author}{Church, K.~W.} (\bibinfo{year}{2022}).
\newblock \bibinfo{title}{Progress in machine translation}.
\newblock {\it \bibinfo{journal}{Engineering}\/},  {\it \bibinfo{volume}{18}\/}, \bibinfo{pages}{143–153}. \URLprefix \url{http://dx.doi.org/10.1016/j.eng.2021.03.023}. \DOIprefix\doi{10.1016/j.eng.2021.03.023}.
\bibitem[{Wang et~al.(2025)Wang, Tang, Guo, Meng, Wang, Wang \& Jia}]{wang2025empowering}
\bibinfo{author}{Wang, X.}, \bibinfo{author}{Tang, Z.}, \bibinfo{author}{Guo, J.}, \bibinfo{author}{Meng, T.}, \bibinfo{author}{Wang, C.}, \bibinfo{author}{Wang, T.}, \& \bibinfo{author}{Jia, W.} (\bibinfo{year}{2025}).
\newblock \bibinfo{title}{Empowering edge intelligence: A comprehensive survey on on-device ai models}.
\newblock {\it \bibinfo{journal}{ACM Computing Surveys}\/}, .
\bibitem[{Yoo \& Cheong(2024)}]{yoo2024leveraging}
\bibinfo{author}{Yoo, T.}, \& \bibinfo{author}{Cheong, Y.-G.} (\bibinfo{year}{2024}).
\newblock \bibinfo{title}{Leveraging llm-constructed graphs for effective goal-driven storytelling}.
\newblock In {\it \bibinfo{booktitle}{CEUR Workshop Proceedings}\/} (pp. \bibinfo{pages}{83--95}).
\newblock \bibinfo{organization}{CEUR-WS} volume \bibinfo{volume}{3818}.

\end{thebibliography}

\end{document}